\documentclass[aps,pre,twocolumn,superscriptaddress]{revtex4-2}  

\usepackage[T1]{fontenc}

\usepackage{amsmath,fixmath}

\usepackage[usenames,dvipsnames]{xcolor}
\definecolor{shadecolor}{gray}{0.9}
\usepackage{lineno}

\usepackage{graphicx}
\usepackage[labelfont=bf,justification=raggedright]{caption}
\usepackage{subcaption}
\usepackage{booktabs}
\usepackage{tabularx}

\usepackage{algorithm}
\usepackage{algpseudocode}
\usepackage{listings}
\usepackage{fancyvrb}
\fvset{fontsize=\normalsize}

\usepackage{natbib}

\usepackage[colorlinks,linktoc=all]{hyperref}
\usepackage[all]{hypcap}
\hypersetup{citecolor=Violet}
\hypersetup{linkcolor=black}
\hypersetup{urlcolor=MidnightBlue}

\usepackage[nameinlink]{cleveref}

\usepackage{comment}

\newcommand{\be}{\begin{equation}}
\newcommand{\ee}{\end{equation}} 
\newcommand{\bea}{\begin{eqnarray}}
\newcommand{\eea}{\end{eqnarray}}

\renewcommand{\bf}[1]{\textbf{#1}} 

\newcommand{\f}[2]{\frac{#1}{#2}}

\newcommand{\ccup}[1]{\left\{#1\right\}}

\newcommand{\bup}[1]{\left(#1\right)}
\newcommand{\rup}[1]{\left[#1\right]}

\newcommand{\id}{\mathbb{I}}

\renewcommand{\ref}[1]{[\ref{#1}]}

\usepackage{amsfonts}
\usepackage{bbm}

\usepackage[textsize=tiny,color=SkyBlue]{todonotes}

\usepackage{titlesec}

\titlespacing\section{0pt}{12pt plus 4pt minus 2pt}{8pt plus 2pt minus 2pt}

\usepackage[normalem]{ulem} 

\newcommand{\nextrout}{\mbox{{\small NextRout}}}

\newcommand{\imgtnet}{\mbox{{\small Image2net}}}
\newcommand{\nefi}{\mbox{{\small NEFI}}}
\newcommand{\mst}{\mbox{{\small Image2net-{MST}}}}
\newcommand{\pp}{\textit{Physarum Polycephalum}}
\newcommand{\river}{\textit{rivers}}
\newcommand{\retina}{\textit{retina}}

\begin{document}

\title{Principled network extraction from images}
\author{Diego Baptista}
\affiliation{ Max Planck Institute for Intelligent Systems, Cyber Valley, Tuebingen, 72076, Germany}
\author{Caterina De Bacco}
\affiliation{ Max Planck Institute for Intelligent Systems, Cyber Valley, Tuebingen, 72076, Germany}


\begin{abstract}
Images of natural systems may represent patterns of network-like structure, which could reveal important information about the topological properties of the underlying subject. However, the image itself does not automatically provide a formal definition of a network in terms of sets of nodes and edges. Instead, this information should be suitably be extracted from the raw image data. Motivated by this, we present a principled model to extract network topologies from images that is scalable and efficient. We map this goal into solving a routing optimization problem where the solution is a network that minimizes an energy function which can be interpreted in terms of an operational and infrastructural cost.  Our method relies on recent results from optimal transport theory and is a principled alternative to standard image-processing techniques that are based on heuristics. We test our model on real images of the retinal  vascular system, slime mold and river networks and compare with routines combining image-processing techniques. Results are tested in terms of a similarity measure related to the amount of information preserved in the extraction. We find that our model finds networks from retina vascular network images that are more similar to hand-labeled ones, while also giving high performance in extracting networks from images of rivers and slime mold for which there is no ground truth available. While there is no unique method that fits all the images the best, our approach performs consistently across datasets, its algorithmic implementation is efficient and can be fully automatized to be run on several datasets with little supervision.
\end{abstract}
\pacs{}

\maketitle

\section{Introduction} \label{introduction}

Extracting network topologies from images is a relevant problem in applications where the subject of the image has a network-like structure. For instance, satellite images of rivers \cite{balister2018river}, neuronal networks \cite{tsai2009correlations,yin2020network}, blood or vein networks \cite{gazit1995scale,boddy2010fungal}, mitochondrial networks \cite{nikolaisen2014automated,ouellet2017novel} or road networks \cite{banavar2000topology}. Assuming this could be done automatically and quantitatively, practitioners would be then able to apply the mathematical study of networks to make quantitative analyses about the topological properties of the system at study. In practice, given a raw image, for instance, a satellite image of a river embedded in a landscape, extracting a network requires identifying a set of nodes and a set of edges connecting them. 
While it might be relatively easy to perform this identification qualitatively, the challenge here is performing this extraction automatically, thus avoiding tedious manual extraction or specific domain knowledge and ad-hoc tools. At the same time, this task should be scalable with system size and number of images as high-quality images are increasingly available and for larger systems.  In addition, a qualitative intuition of the possible existence of a network behind an image is not enough to ensure that no degree of subjectivity is introduced due to the observer's eye. For instance,  two different observers might both perceive the presence of a network-like structure but distinguish two different sets of nodes, and thus two different networks behind the same image. Another challenge is indeed that of performing this extraction in a principled way so that the number of arbitrary choices in defining what the network is should be limited, if not completely absent. 

Here we present a method that addresses these issues by considering the framework of optimal transport. Specifically, inspired by a recently developed model to extract network topologies from solutions of routing optimization problems \cite{baptista2020network}, we adapt this formalism to our specific and different setting. We start from a raw image as input and propose a model that outputs a network representing the topological structure contained in the image. The novelty of this method is that its theoretical underpinning relies on a principled optimization framework. In fact, a proper energy function is efficiently minimized using numerical methods, which results in an output network topology. This implies in particular that network extraction may not depend on the observer's eye, but rather can be automatically done by solving this optimization problem. 
  
  We study our model on real images from different fields, we focus in particular on ecology and biology and compare results with an algorithm that relies on standard image processing techniques, highlighting the main differences resulting from these two different approaches. In particular,  our model allows for an automatic and principled performance of two tasks: filtering network redundancies and selection of edge weights. These are usually challenging tasks for image processing schemes, as they rely on some pre-defined parameter setting in input, while we obtain both directly in output with our model.
   
Many solutions for the problem of automatic network extraction from images have been proposed in computer vision, mainly relying on image-processing techniques \cite{dehkordi2011review,fricker2017automated,lasser2017net,buhler2015phenovein}, for instance segmentation \cite{obara2012bioimage,baumgarten2010detection}, or junction-point processes \cite{chai2013recovering}. The idea is to measure variation of intensity contrast in the image's pixels to highlight curve-like structures.
Within this context, NEFI \cite{dirnberger2015nefi} is a flexible toolbox based on a combination of standard image-processing routines.
 A different approach, closer to the one considered in this work, is that of adopting some sort of optimization framework.  For instance, \cite{breuer2015define} considered an optimization problem where the goal is to minimize the total roughness of a path (a measure depending on the difference of weights in adjacent edges),  in order to decompose a filamentous network into individual filaments. However, these usually rely on domain-specific optimization setups that cannot be easily transferred across domains. Another example is the ant-colony optimization scheme used to extract blood vessels from images of retinas \cite{cinsdikici2009detection}. They all suffer from the NP-hardness of the problem, typical of routing optimization settings. Thus this type of approach relies on approximation techniques. 
Finally, another approach is that of biologically-inspired mathematical models like the one of Tero et al. \cite{tero2007mathematical,tero2010rules} that consider dynamical systems of equations that emulate network adaptability, like that observed for the \textit{Physarum Polycephalum} slime mold. Our model is also inspired by the feedback mechanism of a slime model, which adapts the conductivity of the network edges to respond to differences in fluxes.

Our method relies on the formalism of optimal transport theory used to extract networks from solutions of routing optimization problems proposed in \cite{baptista2020network} and referred to as \nextrout. This is made of three subsequent steps, but here we need only the last two, namely the pre-extraction and the filtering steps. While we refer to that work for all the mathematical details, here we describe the main principles behind this method and adapt it to images. The idea is inspired by the behavior of the \textit{Physarum Policephalum} slime mold. The body of this organism forms a network structure that flexibly adapts to the surrounding environment and the distribution of food sources displaced in it. This network grows with a feedback mechanism between two physical quantities: the conductivities of network edges and the flow passing through them, through a dynamics that is described by a set of equations (sometimes referred to as ``adaptation equations''). In practice, the problem starts by assigning food sources in space and spreading the slime mold uniformly to cover the whole space. The dynamics regulates how the slime mold changes its body shape in time to reach the food in an efficient way. 
The stationary solution of this dynamical system is a set of conductivities and flows on edges that describe the optimal network topology covered by the mold. When the underlying space is continuous, like a squared patch in 2D, these solutions are functions defined on $(x,y)$ coordinates on this space. These are not immediately associated with a network meant as a set of nodes and a set of edges connecting them. However, \cite{baptista2020network} propose principled rules to automatically extract network topologies from these solutions in continuous space. While the main focus of that work was to extract network topologies from this particular type of input (functions defined in a continuous domain, e.g. the space where food sources are located), they hinted at the possibility of adapting this formalism to discrete spaces, like images made of pixels. Here we expand on this insight, and adapt this principled network extraction to inputs that are images. In particular, we propose an algorithm to effectively tackle two problems that are relevant for images and that were only briefly discussed for general applications in \cite{baptista2020network}: how to select source and sinks and how to obtain loopy structures.    \\

\begin{figure}[hptb]
    \centering
    \begin{subfigure}[a]{0.45\textwidth}\label{fig:white}
        \includegraphics[width=\textwidth]{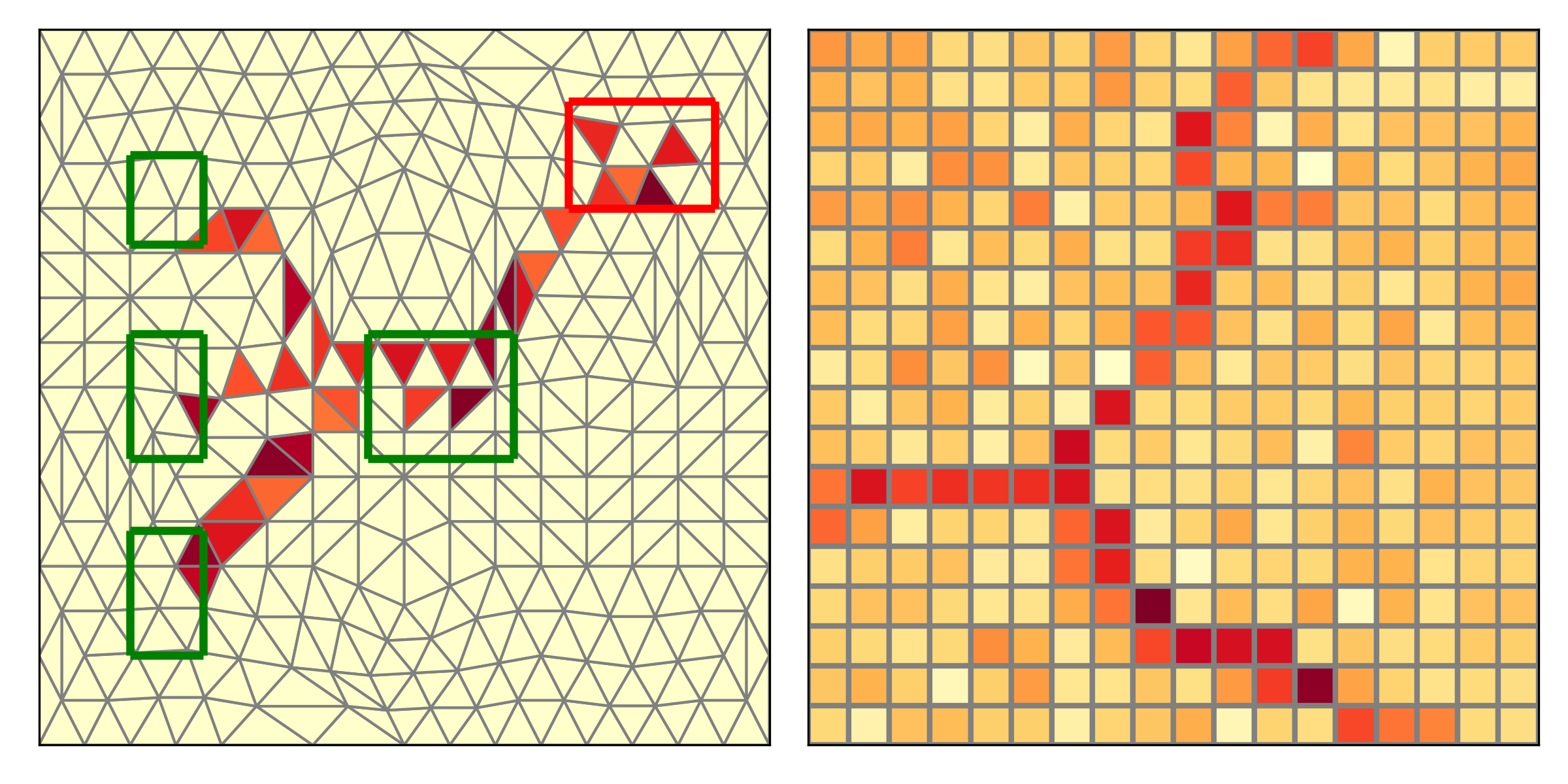}
    \end{subfigure}
      \caption{\textbf{Analogy between optimizing trajectories and images}. Left): the grid structure covering a continuous 2D space and the optimal flux obtained by \nextrout \text{} for a specific routing problem where sources and sinks are inside the green and red rectangles respectively. Red tones of increasing darkness denote higher fluxes on the corresponding grid triangles. Right) the pixel grid and colors of a reference network-like image. }
      \label{fig:2grid}
\end{figure}

\section{ \imgtnet: the method} The key idea is to treat the images as a particular discretization of a 2D space by means of the pixels and treat the RGB color values on them as conductivities. With this setup, we can frame the problem as if there was an imaginary flow of colors that behaves analogously to the slime mold, it starts by covering the whole discrete domain of the image uniformly and then flows through the pixels until it consolidates to a certain subset of them. The observed image corresponds to the network-like shape that the mold converges to in order to optimize the path to reach the food sources. Figure \ref{fig:2grid} illustrates the analogy between the solutions of \nextrout \text{} in continuous space and an image of a network-like structure in discrete space.   

As we mentioned before, this is not yet formally a network, as we do not have a rigorous definition of what constitutes a node and how nodes are connected. However, thanks to the analogy proposed here, we can use the rules introduced in \cite{baptista2020network} for the continuous case and adapt them to images. Specifically, we consider the pixels' barycenters as nodes and draw edges between them depending on their pixels' locations and values, so that two nearby pixels are connected whenever the color has a high enough intensity and their pixels are neighbors. The result is a pre-extracted network that we denote with $G^{pe}$. We denote with  $V$ and $E$ the set of nodes and edges respectively. The network is mathematically encoded by a signed incidence matrix which has entries $B_{ie}=\pm 1$ if the edge $e$ has node $i$ as start/endpoint, 0 otherwise. The sign is important to define the orientation of the flow passing through an edge.  

\begin{figure}[hptb]
    \centering
    \begin{subfigure}[a]{0.45\textwidth}\label{fig:gpe}
        \includegraphics[width=.95\textwidth]{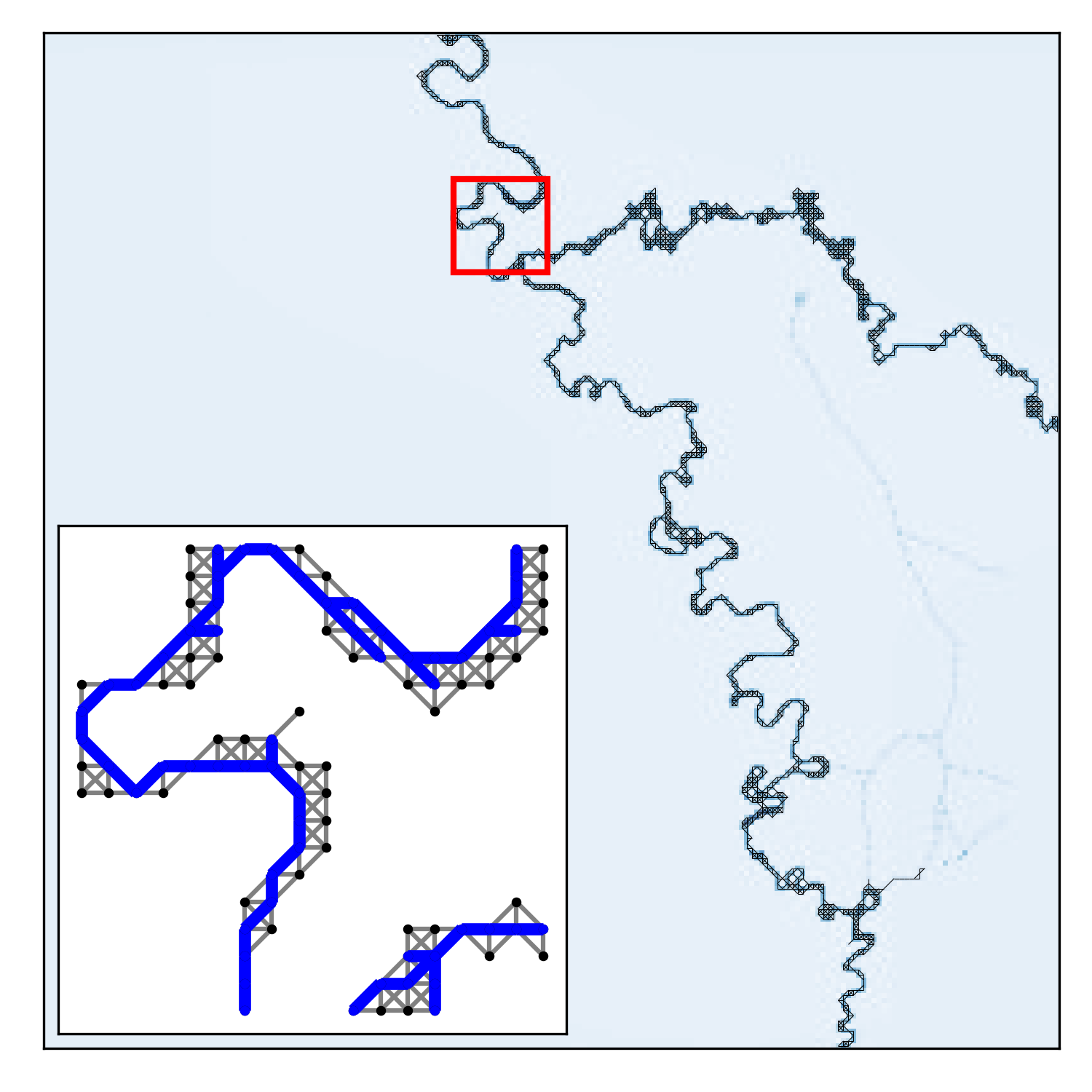}
    \end{subfigure}
      \caption{\textbf{$G^{pe}$ taken from a river image}. The subplot on the bottom left corner shows a section of $G^{pe}$ (in black), highlighted in red in the main plot, together with filtered graph $G^f$ in blue. }
      \label{fig:pre-extracted}
\end{figure}

This temporary network might contain redundancies like dangling nodes or redundant edges, see Fig.~\ref{fig:pre-extracted} for an example. Standard image processing techniques address this problem with pruning routines, e.g. by pruning away edges or branches shorter than a certain length. However, pruning has to be handled with great care, as small redundancies could be a major source of information or they could be completely irrelevant, depending on the network at hand. Usually, pruning is tuned by the user, thus creating potential for subjective bias in extracting the network. Instead, our model relies on a principled method for filtering such redundancies, which exploits a dynamics similar to that of the original problem but adapted to a discrete space like that of the network $G^{pe}$. However, to apply the filter, one must specify a set of terminals, sources and sinks, as input to the discrete dynamics. Continuing with our analogy, we need to locate the pixels where we imagine that color mass is being injected and extracted. These are the sources and sinks that drive the dynamics to consolidate the flux of colors on the network-like structure observed in the input image.  

\subsection{Dynamics}\label{sec:dyn}
Assume for a moment that we knew this set of terminal pixels and denote with $f_{i}$ the amount of color mass that enters or exits the image in node $i$ (the barycenter of pixel $i$). Note that to preserve the mass, we have $\sum_{i}f_{i}=0$. Here we describe in more detail the dynamical rules that regulate how colors spread along the pixels in an optimal way.
To describe the flow of colors, we consider the conductivity $\mu_{e}$ on an edge $e \in E$ and the potential $u_{i}$ on a node $i\in V$. The conductivities can be interpreted as the size of the diameter of an edge, the potentials as pressures on nodes. Together, these two quantities determine the flow $F_{e}$ of color passing through an edge $e=(v_{1},v_{2})$ in the network:

\bea
F_{e} &=& \f{\mu_{e}}{\ell_{e}} \bup{u_{v_{1}}-u_{v_{2}}} = \f{\mu_{e}}{\ell_{e}}\left\lvert\sum_{j\in V}B_{ej}\,u_{j}\right \rvert \quad.
\eea

 In turns, the flow influences the conductivities and potentials, through a feedback mechanism described by the following set of equations:
\bea\label{eqn:discreteDMK1}
 f_{i}  &=&\sum_{e\in E} B_{ie} F_{e} \,, \label{eqn:kirkdiscrete}\\ 
 \mu_{e}'(t)&=&\rup{\f{\mu_{e}(t)}{\ell_{e}}\left\lvert\sum_{j\in V}B_{ej}\,u_{j}(t)\right\rvert}^{\beta}-\mu_{e}(t)  \label{eqn:mudiscrete}\,,\\ 
 \mu_{e} (0)&>&0 \,,\label{eqn:ICdiscrete} 
\eea
where $|\cdot|$ is the absolute value, $\ell_{e}$ is the euclidean length of an edge using the barycenters' coordinates and $\beta$ is a parameter that determines the optimization mechanism.
Equation (\ref{eqn:kirkdiscrete}) is Kirchoff's law, Eq. (\ref{eqn:mudiscrete}) is the discrete dynamics describing the feedback mechanism between conductivity and flow: when the flow of colors is high on an edge $e$, the conductivity increases, and vice versa when the flow is low the conductivity decreases;  Eq. (\ref{eqn:ICdiscrete}) is the initial condition. 
The stationary solution of this dynamical system can be mapped to the solutions of an optimization problem where the cost function can be interpreted as a network transportation cost:
\bea
\mathcal{L}_{\beta}(\mu(t)) &=&\f{1}{2} \sum_{e}\mu_{e}(t) \bup{\f{1}{\ell_{e}}\sum_{j}B_{ej}u_{j}(\mu(t))}^{2}\ell_{e} \nonumber\\
&&+\f{\beta}{2}\sum_{e}\f{\mu_{e}(t)^{(2-\beta)/\beta}}{2-\beta}\, \ell_{e} \label{eqn:lyap} \quad,
\eea
where $\mu(t) = \ccup{\mu_e(t)}_e$, and the first term is the network operational cost and the second is the cost to build the network. The values of $\mu_{e}$ at convergence can be used not only to determine the set of edges in the extracted network,  but also its weights, which can be interpreted as the diameter of the edge on the image. This is one of the advantages of our model, as estimating the diameter of edges extracted from an image is an open problem when using image-processing techniques. We get this automatically with the optimal conductivities.

The dynamics works as a filter, i.e. removes redundancies, for $\beta \geq1$. In this work, we fix $\beta=1.5$ as it gives good performance consistently across the datasets studied here.  The output result is a tree, i.e. it does not contain loops and is the optimal one in terms of minimizing the transportation cost of Eq.~(\ref{eqn:lyap}).
In our experiments, we use the numerical solver proposed in \cite{baptista2020network} to extract the stationary solutions of the system of Eq.~(\ref{eqn:kirkdiscrete})-(\ref{eqn:ICdiscrete}). 

\subsection{Selecting terminal pixels}\label{sec:sourcesink}
Having introduced how the dynamics of the colors works, we now tackle the problem of selecting the terminal pixels where to inject or extract imaginary color mass. This choice is crucial as it determines the final extracted networks, in the same way, that the location of the food sources determines the particular shape of the network that a slime mold will form to optimally reach them in a 2D space.   In the original problem of  \cite{baptista2020network}, this was not an issue because the set of terminals could be selected from that of the original problem in continuous space. In other words, this was an input of the problem. Here we do not start from that input, but rather have access to only a raw image, without any notion of pre-specified terminals attached to it. In practice, we need to find the pixel nodes corresponding to the rectangles inside Fig.~\ref{fig:2grid} (left). Here we propose a method to make this selection effectively.
Specifically, we select as a set of eligible terminals $\mathcal{T}(G^{pe}_{tree})$ all the leaves of the tree $G^{pe}_{tree}$ obtained from running our dynamics on $G^{pe}$ when we pass in input all pixel nodes in $G^{pe}$ as terminal, and selecting one of these at random as a source, all the rest as sinks. 
This choice is motivated by the fact that the tree resulting from the filtering is a good approximation of the pre-extracted  $G^{pe}$, as it follows a principled optimization framework. The obtained leaves determine the coverage of this network, as they are usually located in distant parts of the network.
Potentially, one could select terminal pixels ``manually'', by using domain-knowledge to determine what pixels are the most important.  However, this strategy is not scalable to a large number of images. Instead, our proposed procedure does not suffer from this problem as it can be automatically implemented, while also being flexible to receive ``hand-picked'' terminals if available. Alternatively, a practitioner could make this selection based on some notion of network centrality, for instance selecting as terminal the most ``central'' nodes. However, this again assumes having extra information to decide what definition of centrality is appropriate based on the application. We do not explore this here.

\subsection{Obtaining loops}
Running the dynamics of Sec. \ref{sec:dyn} outputs trees, while network-like structures in images might have loops. The question is thus how to recover networks that are not limited to trees. 
We tackle this problem by re-running the dynamics multiple times, each time selecting a particular node as source from the eligible ones (and sinks all the others). Specifically, we randomly select an individual source pixel $i \in \mathcal{T}(G^{pe}_{tree})$ and assign all the others $j \neq i \in \mathcal{T}(G^{pe}_{tree})$ as sinks. 
Applying the dynamics to $G^{pe}$ with this choice of one source and multiple sinks outputs a filtered network $G_{r}^{f}$, indexed by the iteration run $r$. By repeating for $N_{runs}$ this filtering step, each time selecting a different source from $\mathcal{T}(G^{pe}_{tree})$ (and all the remaining node pixels as sinks), results in a set of filtered networks $\ccup{G_{1}^{f},\dots,G_{N_{runs}}^{f}}$, all of them trees. 
We combine them by superposition, so that we obtain a final network $G(V,E,W)$ where the set of nodes and edges are the unions $V = \bigcup_{r=1}^{N_{runs}} \! \! V(G^f_r)$, and $E = \bigcup_{r=1}^{N_{runs}} \! \! E(G^f_r)$. The weights on the edges of the final network are given by the sum of the weights on each run
\be
w_{jk} = \sum_{r=1}^{N_{runs}}w^{r}_{jk}, \ \ \forall j,k \in V,
\ee
where $w^{r}_{jk}$ is the weight of edge $(j,k)$ in network $G^{f}_{r}$ and corresponds to the optimal edge conductivities as obtained from the dynamics at convergence. We assume $w^r_{jk}=0,$ if $(j,k)\not \in G^f_r.$
The value of $N_{runs}\leq |\mathcal{T}(G^{pe}_{tree})|$ is a parameter that has to be tuned based on the input image. Notice that a high value of $N_{runs}$ might not necessarily result to a network more similar to the one depicted in the input image. For instance, in the extreme scenario where the original network-like structure is a tree, then $N_{runs}=1$. Empirically, we find that a value of $N_{runs} = 5$  gives good results in all the experiments reported here, see Supplementary Information for more details.

Combining these steps we obtain the whole algorithmic pipeline of our method, which we refer to as \imgtnet. We provide an algorithmic pseudo-code in Algorithm \ref{alg:pipeline} and an open-source implementation at \url{https://github.com/diegoabt/Img2net}.

\begin{algorithm}[H]
   \caption{\imgtnet}
   \label{alg:pipeline}
\begin{algorithmic}[1]
\Require{Image $\mathcal{I}$, threshold $\delta$, $\beta \geq1$}
\Ensure{$G(V,E,W)$ final network}
    \Function{\imgtnet}{$\mathcal{I},\delta,\beta\geq1$}
 \State $ G^{pe} \gets $  \nextrout \text{} pre-extraction($\mathcal{I},\delta$)
    \State $G^{pe}_{tree} \gets $ run Dynamics of Sec. \ref{sec:dyn} on $G^{pe}$ \Comment{ for $\beta$ and using in input as starting sources and sinks all the nodes in $G^{pe}$}
   \State  $\mathcal{T}(G^{pe}_{tree}) \gets \ccup{v \in V(G^{pe}_{tree}) | d_{v}=1} $  \Comment{$d_{v} $ is the degree of node $v$}
     \For{\texttt{$r = 1, \dots ,N_{runs}$}}
        	\State {Select $i \in \mathcal{T}(G^{pe}_{tree})$ \Comment{uniformly at random}}
	        	\State {$G_{r}^{f} \gets$ run Dynamics of Sec. \ref{sec:dyn} on $G^{pe}$ \Comment{ for $\beta$ and using as starting source $i$ and sinks $\ccup{j \neq i \in \mathcal{T}(G^{pe}_{tree})}$ }}
      \EndFor
 \State $G(V,E,W) \gets $ Superimpose$\ccup{G_{1}^{f},\dots,G_{N_{runs}}^{f}}$
   \EndFunction
  \end{algorithmic}
\end{algorithm}

\section{Experiments on images}\label{exp-setup}
We run our model on three datasets of images covering various types of network-like topologies observed in biology and ecology. The images represent: i) the slime mold Physarum Polycephalum (\pp) \cite{dirnberger2017introducing}, which is also the inspiration of our dynamics; ii) the retinal vascular system (\retina) \cite{hoover2000locatingbv}; iii) river networks (\river) obtained by extracting images from \cite{openseamap}. 
The number of images taken from the \pp, \retina  {} and \river  {} sources is 25, 20 and 10 respectively, see Table~\ref{tab:data_desc}. Pre-processing was applied, see Supplementary Information for details. 

\begin{table}[htbp]
\begin{center}
\caption{{\bf {Datasets description.} NI is the number of images used; AW is the average width of the images in the dataset; MinW and MaxW denote respectively the minimum and maximum width of the images in the dataset.}}
\resizebox{1\columnwidth}{!}{%
{\renewcommand{\arraystretch}{1.11}
\begin{tabular}{llllll}
\toprule
 \textbf{Dataset} & \textbf{Description}  &\textbf{NI}  & \textbf{AW} & \textbf{(MinW,MaxW)}  & \textbf{Ref.}\\
\midrule
\retina & Retinal blood vessels & 20 &$1791$&$(998,2302)$  &  \cite{hoover2000locatingbv,stare}  \\
\pp & Slime mold&25 &$400$&$(400,400)$&   \cite{dirnberger2017introducing}  \\
\river& Riverbed &10 &$924$ &$(718,958)$& \cite{openseamap}    \\
\bottomrule
\end{tabular}%
}}
\label{tab:data_desc}
\end{center}
\end{table}

For model comparison, we consider NEFI, a routine that combines various image processing techniques, and a variant of our routine based on a combination of Minimum Spanning Tree and Steiner Tree optimization (\mst). The idea behind this last routine is to run our procedure but replacing the optimization steps based on the dynamics of Sec. \ref{sec:dyn} with standard routing optimization algorithms, namely a combination of Minimum Spanning and Steiner trees \cite{hwang1992steiner}. The goal is to see how the underlying optimization setup impacts the final network topology. In fact, while the underlying idea of treating the problem of network extraction from images within the framework of routing optimization is the same for \imgtnet \text{} and \mst, the details of their corresponding optimization differ.  
Specifically, for \mst, we first run a standard MST optimization algorithm to extract $G^{MST}_{tree}$ from $G^{pe}$ (this is the same input given to \imgtnet). From the corresponding set of leaves $\mathcal{T}(G^{MST}_{tree})$, this time one should extract a subset of terminals $T\subseteq \mathcal{T}(G^{MST}_{tree})$ of predefined size (no distinction between source and sinks is necessary to solve a minimum Steiner tree problem). From $G^{pe}$ and $T$, extract a Steiner tree $G_{r}^{St}$, repeat this $N_{runs}$ times and obtain the set $\ccup{G_{1}^{St},\dots,G_{N_{runs}}^{St}}$. Finally, superimpose them as done for our method to obtain $G_{mst}$, see Supplementary Information for more details. Notice that Steiner tree optimization has a complexity that scales with the number of terminals, a problem not present in our dynamics. As a result, running \mst \text{} is noticeably computationally more expensive than \imgtnet. 

\begin{figure*}[hptb]
    \centering
    \begin{subfigure}[a]{0.99\textwidth}\label{fig:white}
        \includegraphics[width=\textwidth]{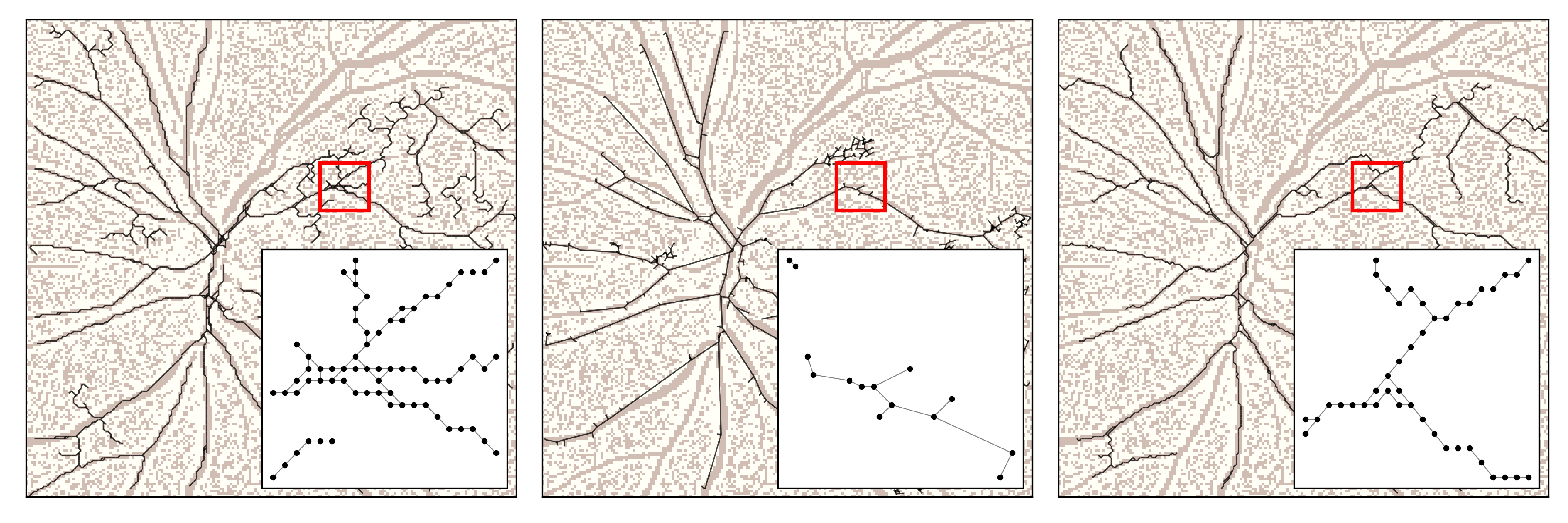}
    \end{subfigure}
      \caption{\textbf{Example of network extraction}. {Left) \imgtnet; center) \nefi; right) \mst.} The extracted network is colored in black, the original image is in light brown underneath. The inset is a zoom inside the image section highlighted in red.}
      \label{fig:3meth}
\end{figure*}

Finally, edge weights were assigned with rules specific to each method, as there is no common definition that applies to all of them. In fact, the ability to extract edge weights is rare among image processing techniques,  and usually relies  on image preprocessing and segmentation of the input image. Instead,
\imgtnet \text{} extracts edge weights in a principled way based on the results at optimality in terms of conductivity, hence it has a nice direct interpretation as the diameter of the edges in the image. For \imgtnet, we use the rule Effective Reweighing (ER) on the resulting conductivities, see \cite{baptista2020network}; for \mst, we use the weights given  in input to solve the Steiner tree problem, i.e. the weights given by ER rule in $G^{pe}$; for \nefi, we use as weight the \textit{width}, this is an output of the algorithm; for the original image we assign the RGB values of the pixels mapped into an integer number increasing with the color intensity. All of these definitions of weight agree on the higher the weight the thicker the edge is, and thus the conductivity.  
Figure \ref{fig:3meth} illustrates an example of the networks extracted using the various algorithms for an image in \retina. 

\subsection{Performance metrics}\label{sec:metrics}
We measure performance in terms of the ability of an algorithm to recover the network-like subject depicted on the underlying image. We consider a measure of similarity adapted from the quality measure defined in \cite{baptista2020network}.  This relies on partitioning the image in a grid of $P$ non-intersecting subsets $C_{\alpha}$ inside the pixels' domain and then compare the edges of $G$ within $C_{\alpha}$ assigned by the algorithm and those observed in the original image $I$ (RGB values):
\be\label{eqn:bin_qm}
\hat{w}_b(G,I) = \f{1}{P}\rup{\sum_{\alpha=1}^{P} \bup{|\sum_{e \in E} \id_{\alpha}(e) -\sum_{i\in I}\id_{\alpha}(\delta,i)|}^{2}}^{1/2}\quad,
\ee
where $\id_{\alpha}(\delta,i)= 1,0$ for $i\in I$, if the pixel $i$ is in $C_\alpha$ or not, respectively; $\delta$ is  a threshold used to decide whether that pixel contributes to the network-like image. In words, if the pixel color intensity is high enough, then we label it as an edge. For $\delta$, we use the same value as used in input to \imgtnet.
This is a coarse-grained measure of similarity that tells how many edges in the extracted graph correspond to high-intensity pairs of pixels.
In order to account for edge weights and pixel intensities, we also consider  a weighted version of this: 

\be\label{eqn:qm}
\hat{w}(G,I)= \f{1}{P}\rup{\sum_{\alpha=1}^{P} \bup{|\sum_{e \in E} \id_{\alpha}(e) w_{e}-\sum_{i\in I}\id_{\alpha}(i)p_i|}^{2}}^{1/2}\quad,
\ee
where $p_i$ is the intensity of the pixel $i$, and $\id_{\alpha}(i)$ is 1 if $i \in C_\alpha, $ and 0 otherwise. Notice that in this case $\delta$ is not needed since  pixels with low intensity are penalized by lower weight in their contributions to $\hat{w}(G,I)$. In both cases, small values of these measures mean higher similarity values between the extracted network and the underlying network-like structure in the image.


While ground-truth for this network-like structure is normally absent, the \retina \text{} dataset contains ground-truth networks which were hand-labeled by individuals \cite{hoover2000locatingbv}. In this case, we calculate the binary similarity using the hand-labeled images instead of the one given in input. There are two sets of labeled images, each corresponding to a different person doing this manual identification. While similar, the resulting two sets of networks are different. In the absence of ground truth, we compare against the input image. 

\subsection{Implementation details.}   
We apply image pre-processing to the input image to improve image quality and distinguish the main subject from the background, see Supplementary Information for details.  We rescale \nefi's pixels' location to have them in the same scale as that of the other methods, i.e. the set $[0,1]\times [0,1]$ (for simplicity, we consider only square-shaped images). All the edge lengths $\ell_{e}$ have been assigned using the Euclidean distance between the corresponding endpoints. For \nefi, we used the two best performing pipelines of image-processing techniques \textit{polycephalum\_high} (\texttt{NEFI-high}) and \textit{crack\_patterns} (\texttt{NEFI-crack}) among the available predefined pipelines. In the figures we show the best results only, these vary based on the image given in input.

\section{Results}
\subsection{Retinal vessel image validation} 
\begin{figure*}[hptb]
    \centering
    \begin{subfigure}[a]{0.4\textwidth}
        \includegraphics[width=1\textwidth]{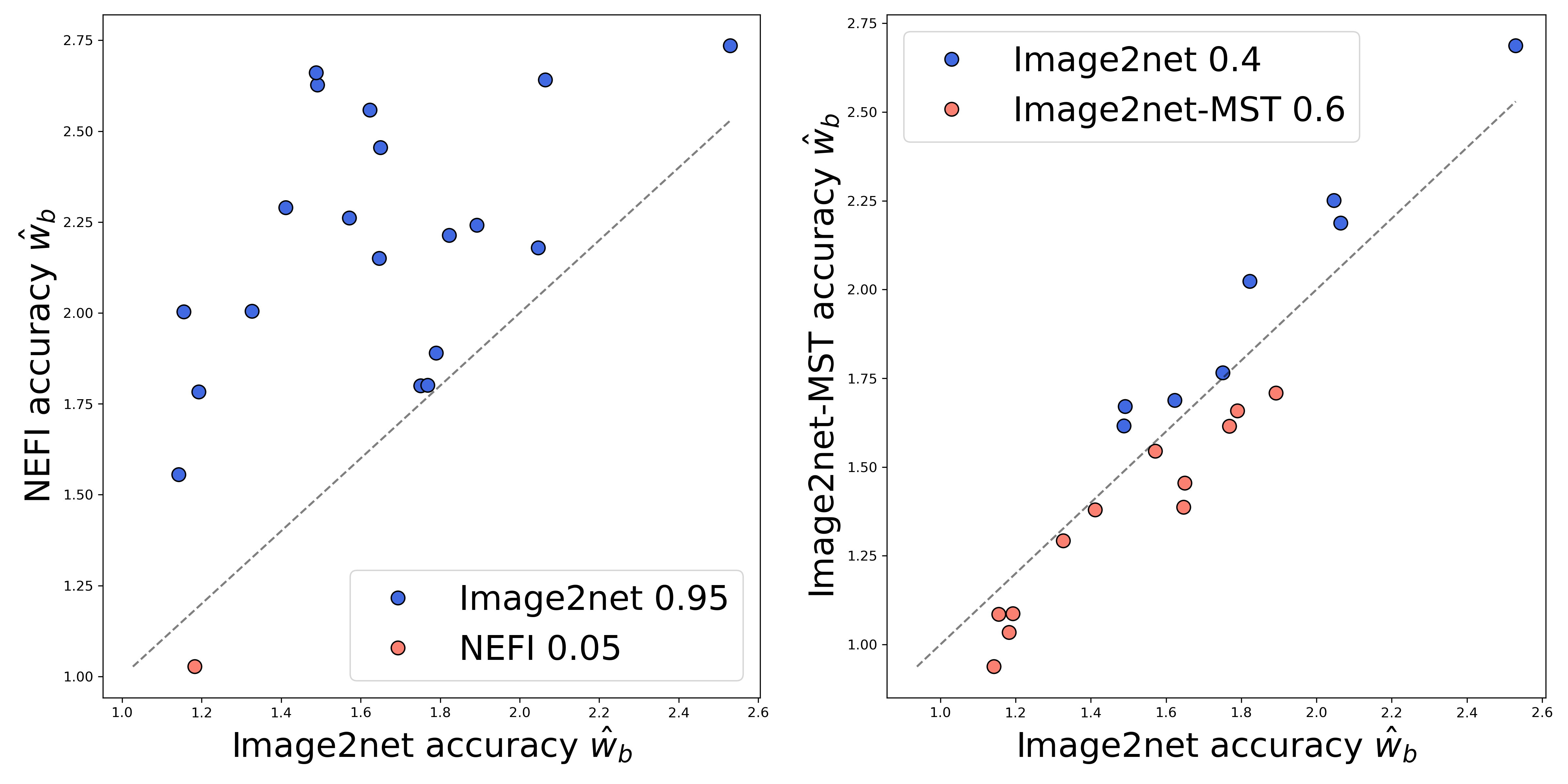}
         \includegraphics[width=1\textwidth]{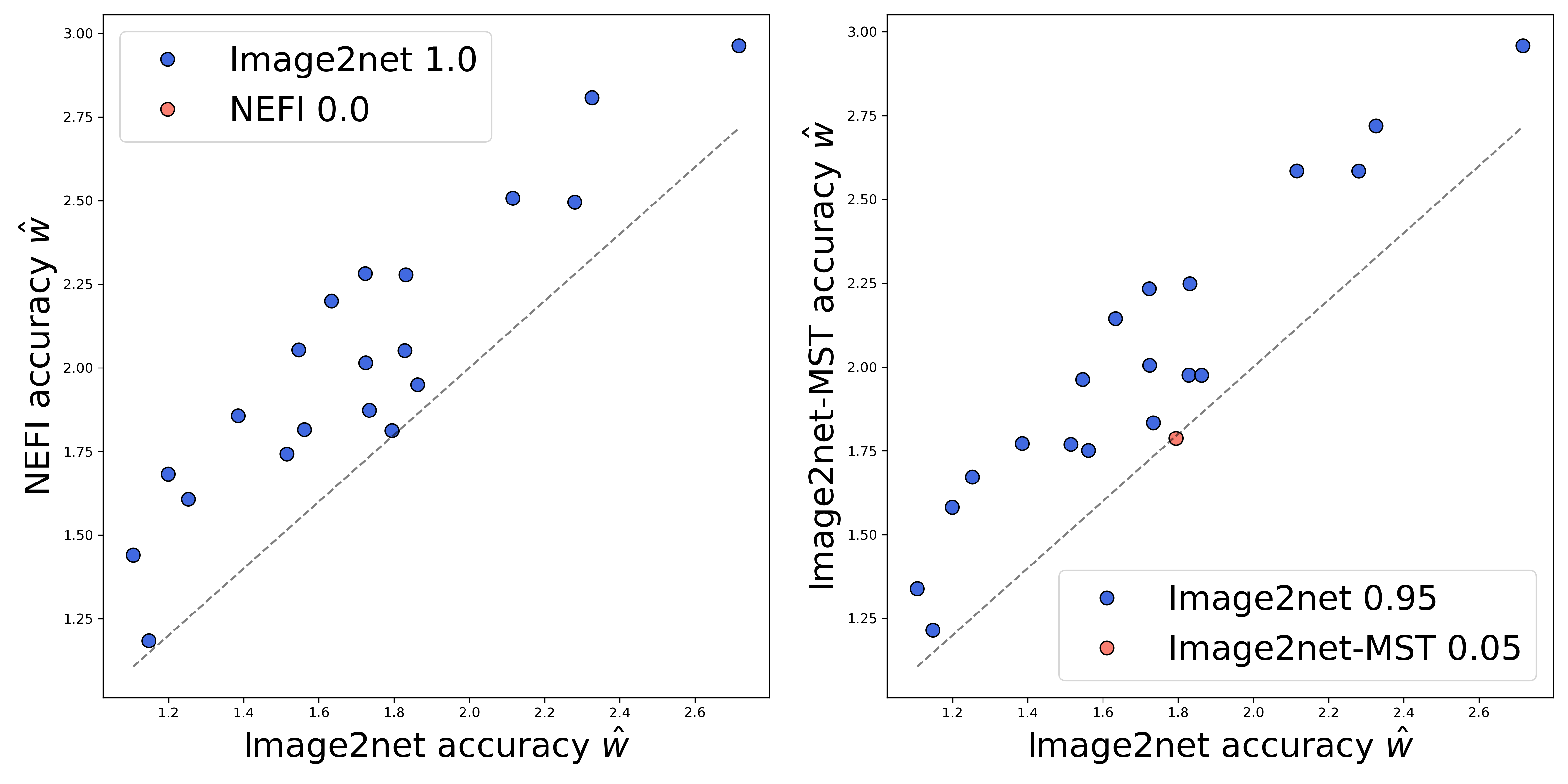}
         \caption{Hand-labeled by A. H.}
    \end{subfigure}
        \begin{subfigure}[a]{0.4\textwidth}
        \includegraphics[width=1\textwidth]{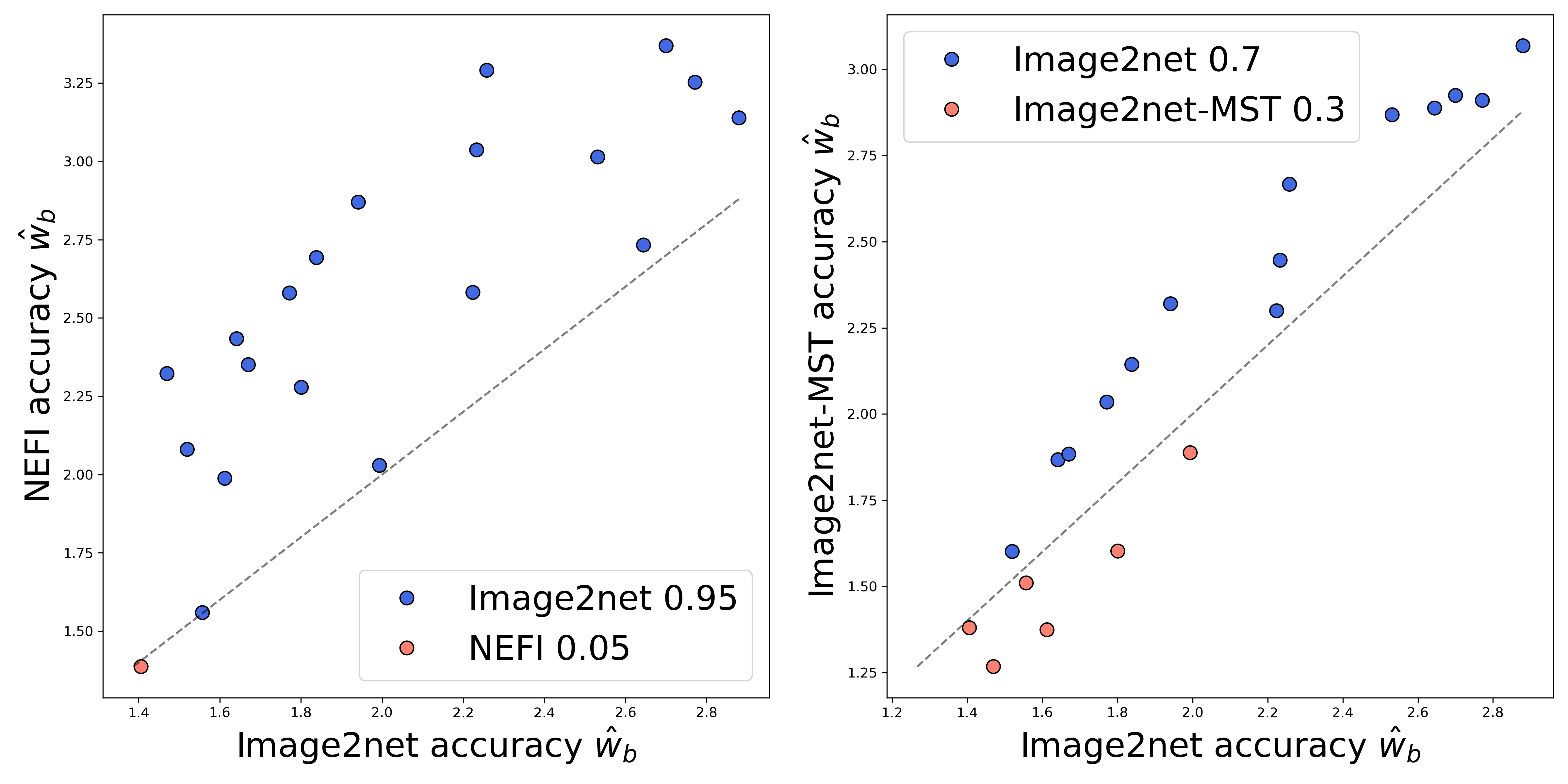}
          \includegraphics[width=1\textwidth]{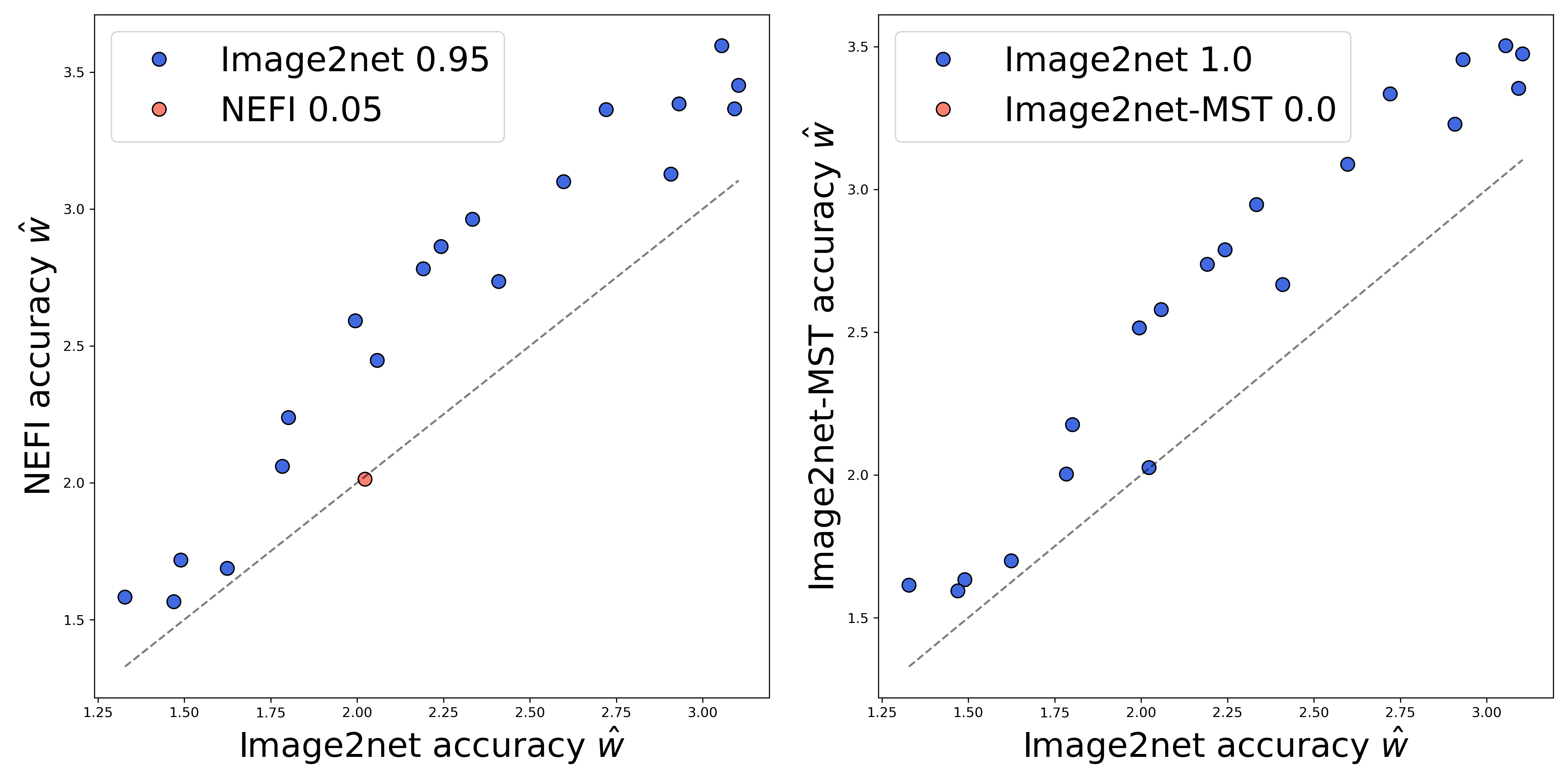}
          \caption{Hand-labeled by V. K.}
    \end{subfigure}
      \caption{\textbf{Recovering hand-labeled networks}. Performance in terms of similarity $\hat{w}_b(G,I)$ and $\hat{w}(G,I)$ on hand-labeled \retina \text{} networks, (a) and (b) are networks labeled by two different people. First row shows $\hat{w}_b$ values; second row shows $\hat{w}$ values. Smaller values mean higher similarity and thus better performance. Hence, points above the grey line (blue) means \imgtnet \text{} performs better, whereas points below (red) means worse performance. }
      \label{fig:qm-retinal-hand-labeled}
\end{figure*}

We use the similarity measure defined in the previous section to compare every graph-based approximation of the image with the provided hand-labeled ones, assuming these last ones to be the ground truth $I_{gt}$. We compute $\hat{w}_b(G,I_{gt})$  for each retinal image and the corresponding extracted network, to measure how close a particular network is from the human-labeled one.
Fig.~\ref{fig:qm-retinal-hand-labeled} shows that \imgtnet \text{} consistently outperforms \nefi \text{} over all images and the two hand-labeled datasets. \imgtnet \text{} and \mst \text{} perform similarly according to the binary similarity. However, if we account for weight, we obtain the \imgtnet \text{} outperforms \mst \text{} in the majority of the images. Notice that \mst \text{} does not assign new weights while selecting the edges,  as in a Steiner tree problem, instead, it uses the weights of the input network, in this case, $G^{pe}$. Instead, \imgtnet \text{}  selects edges and weights at the same time, within the same optimization setup. The fact that the weighted similarity gives better results,  signals that the values of the optimal conductivities (the weights assigned to \imgtnet \text{} extracted networks) have a meaningful interpretation, as they better match the pixel's intensities than the weights given by the other algorithms.


\subsection{Physarum Polycephalum and river networks.}
\begin{figure*}[hptb]
    \centering
        \begin{subfigure}[]{0.4\linewidth}
         \centering
        \includegraphics[width=1\linewidth]{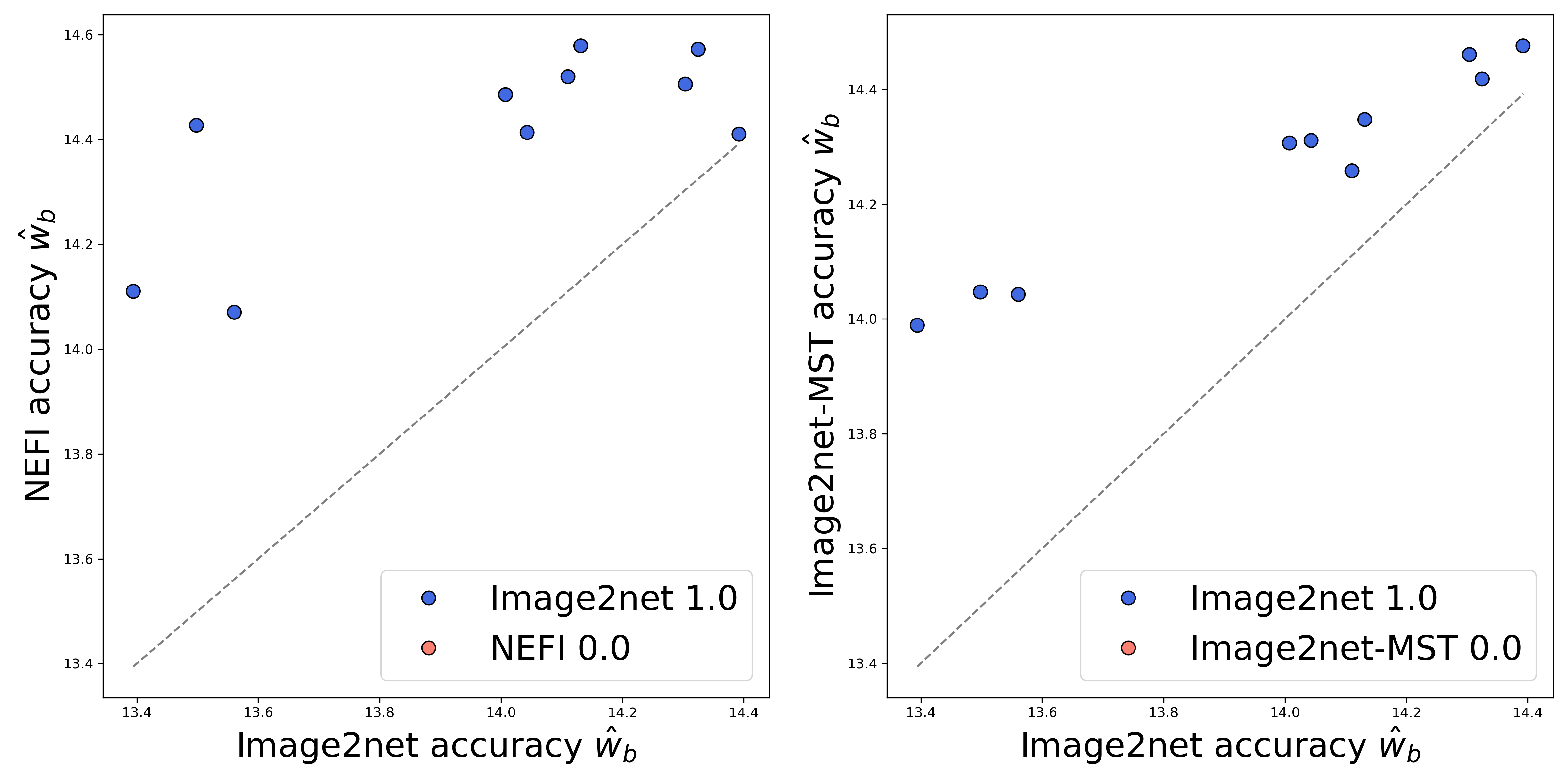}\\
         \includegraphics[width=1\linewidth]{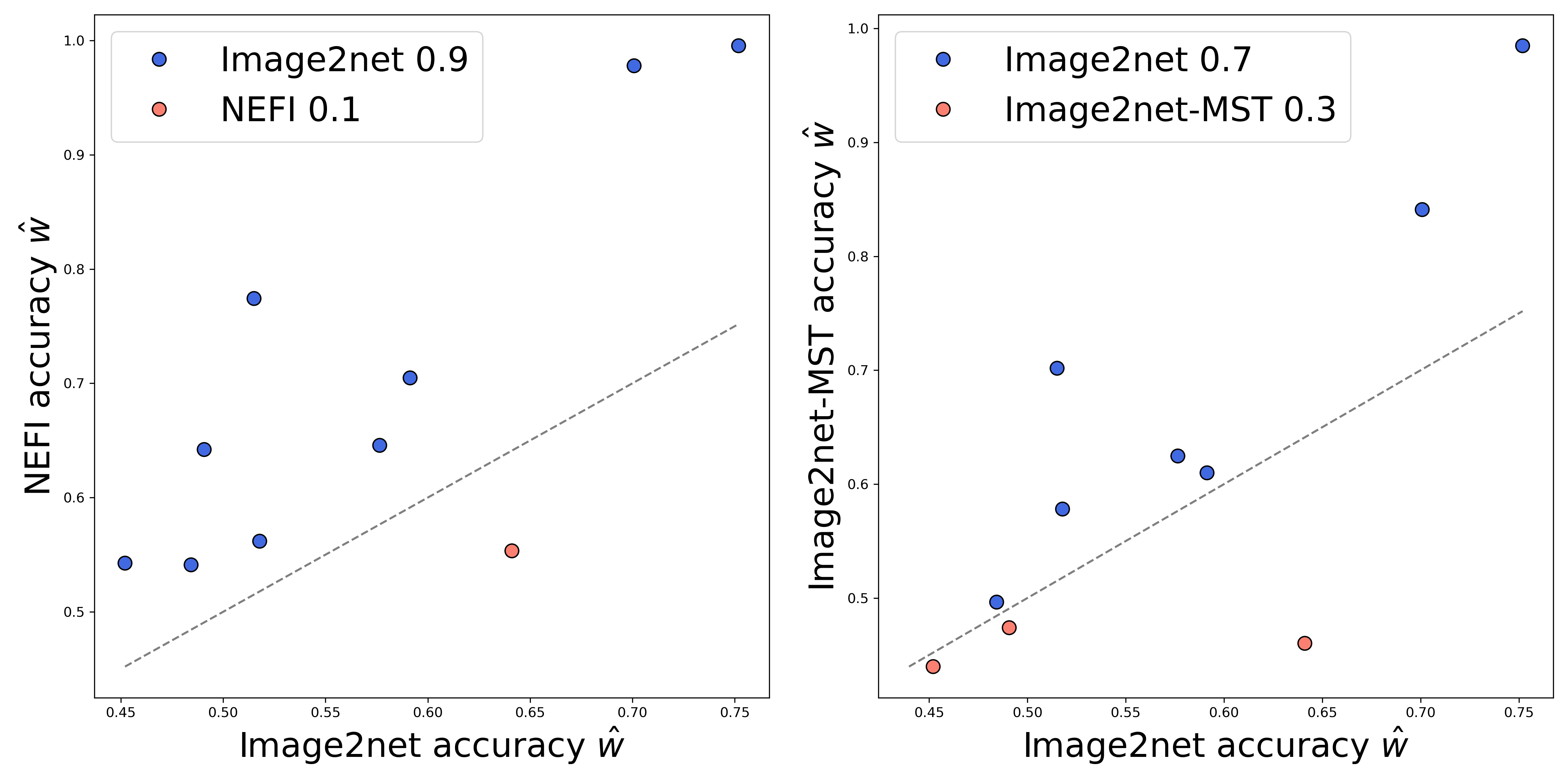}
             \caption{\river}
    \end{subfigure}%
        \begin{subfigure}[]{0.4\linewidth}
        \centering
        \includegraphics[width=1\linewidth]{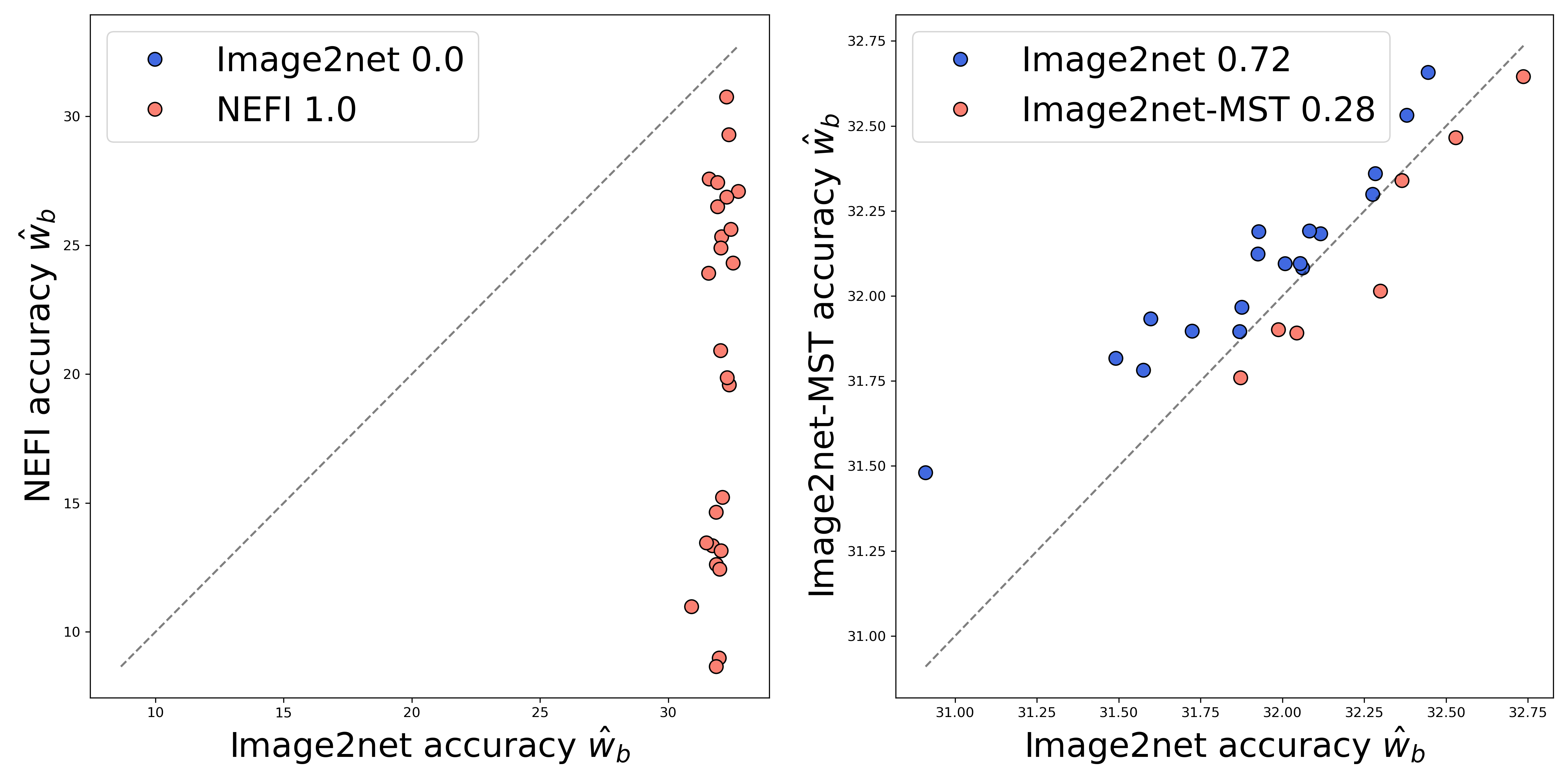}\\
          \includegraphics[width=1\linewidth]{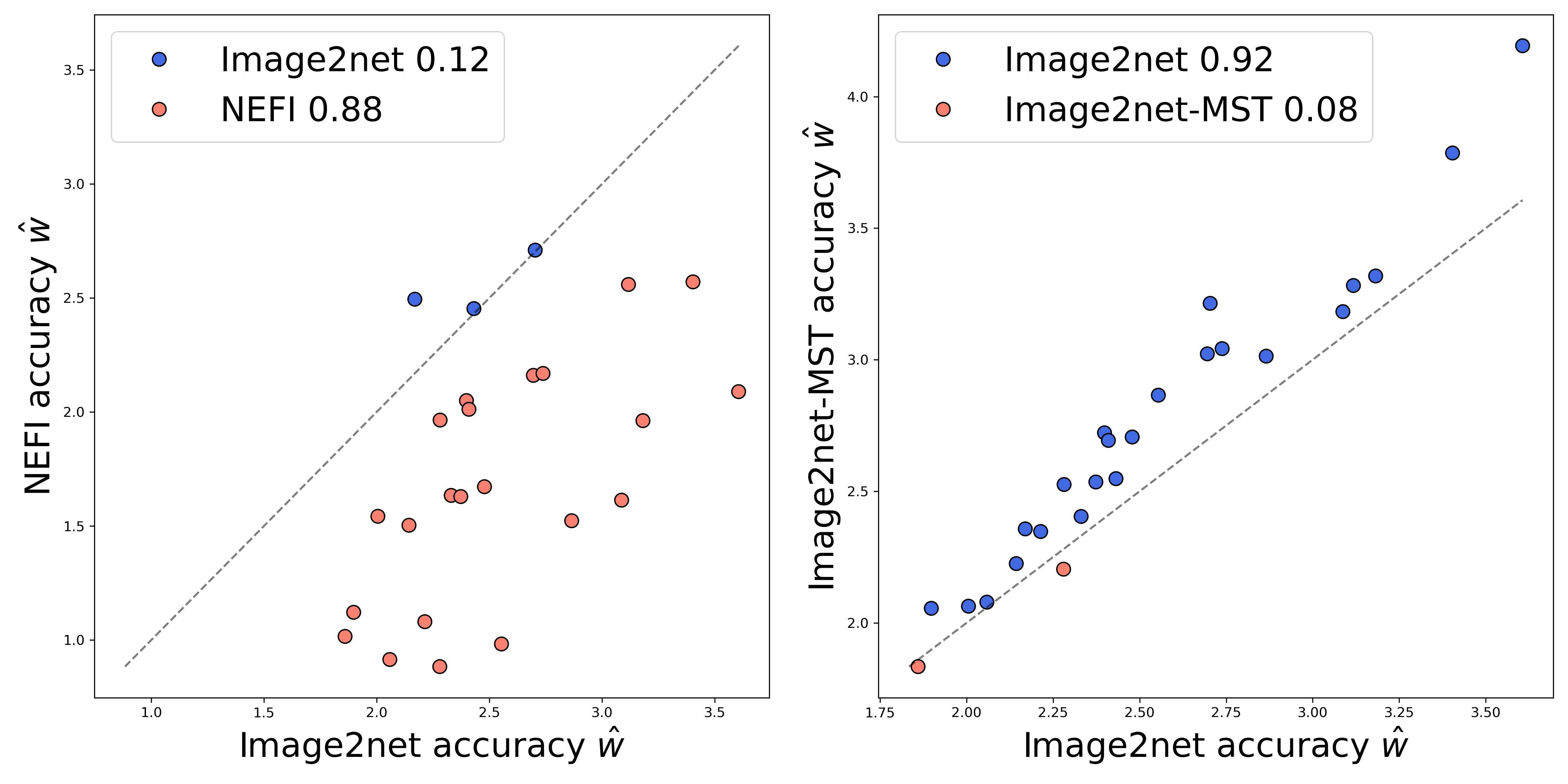}
          \caption{\pp}
    \end{subfigure}%
      \caption{\textbf{Recovering river and physarum polycephalum networks}. Performance in terms of similarity $\hat{w}_b(G,I)$ and $\hat{w}(G,I)$ on \river \text{} (2 leftmost columns) and \pp \text{} (two rightmost columns) networks. Smaller values mean higher similarity and thus better performance. Hence, points above the grey line (blue) means \imgtnet \text{} performs better, whereas points below (red) means worse performance.}
      \label{fig:qm-no-ground-truth}
\end{figure*}

\begin{figure*}[tb]
    \centering
        \includegraphics[width=0.9\textwidth]{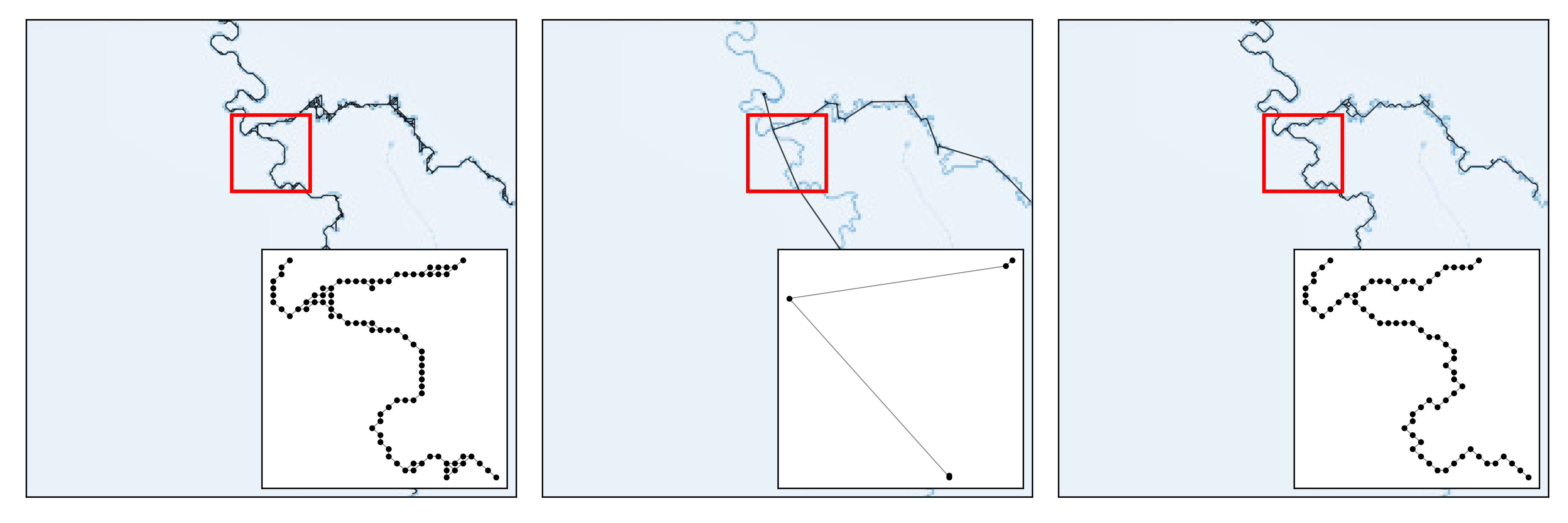}
         \includegraphics[width=0.9\textwidth]{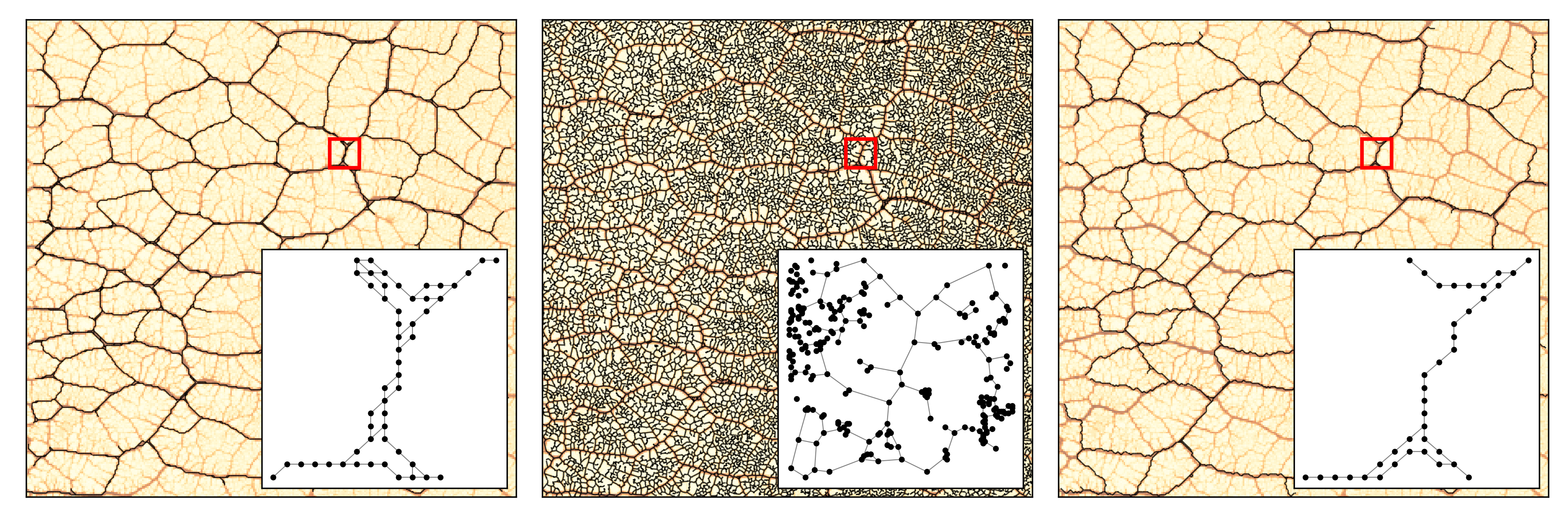}
      \caption{\textbf{Results on \river \text{} and \pp \text{} networks}. We show the networks extracted on rivers (top) and PP (bottom) using \imgtnet \text{}(left), \nefi \text{} (center) and \mst \text{} (right). Inset is the zoom over the area under the red surface. The input image is depicted underneath the networks.}
      \label{fig:networks-rivers-pp}
\end{figure*}

We measure the performance in the two datasets where there is no ground truth, which is often the case in real images. We find that \imgtnet \text{} recovers better the \river \text{} networks, for both performance metrics as we show in Fig.~\ref{fig:qm-no-ground-truth}. In fact, our model is able to capture the detailed geometry of the network when there are curves,  while \nefi \text{} has limitations in that edges with curves or kinks are contracted to straight lines. This is one of the main advantages of our model based on an underlying optimization framework, the geometry of the network is automatically selected based on optimality, rather than a predefined setting manually tuned. As a result, \imgtnet \text{} is flexible in detecting different network geometries, as can be seen in Fig.~\ref{fig:networks-rivers-pp} (top). The situation for \pp \text{} is more nuanced as \imgtnet \text{} is better than \mst, in particular when considering the weights, but \nefi \text{} outperforms all the others. However, this is true if we use the \texttt{NEFI-high} routine, which is the one built on purpose to detect \pp \text{} networks, it is not surprising that this has stronger results on these datasets. In Fig.~\ref{fig:networks-rivers-pp} (bottom) we notice how these networks contain many small details that are better captured by \nefi. Indeed, using other \nefi \text{} routines, performance aligns more to \imgtnet \text{} and \mst. This also shows that if a practitioner aims at extracting networks from a particular image, all the approaches allow for few degrees of freedom to be tuned in order to increase performance. \nefi \text{} allows to specify individual routines to design a custom pipeline, \imgtnet \text{} and \mst \text{} have various parameters that could be tuned, the most important being $\delta$ and $\beta$. For instance, decreasing $\delta$ will allow for more fine details on \pp \text{} networks, see Supplementary Information. However, tuning each routine on each input image  goes beyond the scope of this work, as we aim at describing how different approaches perform on a corpus of images, potentially quite different, and thus automatize network extraction in a scalable way.


\section{Qualitative results}
Beyond validating the model on recovering network structure that resembles well what is pictured in an image, we illustrate the differences in topological properties of the extracted networks. This also showcases possible applications for our model, where a practitioner extracts a network and can then perform further analysis on it, for instance using the detected network properties. 

\begin{figure*}[hptb]
    \centering
         \begin{subfigure}[a]{0.32\textwidth}
         \includegraphics[width=1\textwidth]{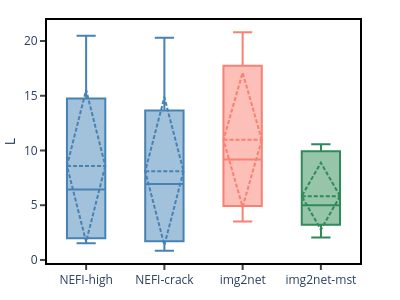}
         \caption{\river}
             \end{subfigure}
              \begin{subfigure}[a]{0.32\textwidth}
           	\includegraphics[width=1\textwidth]{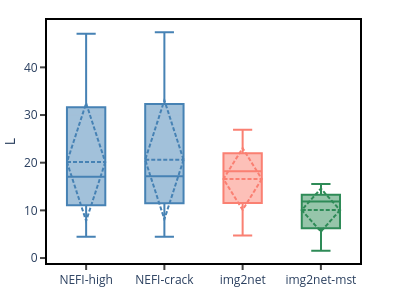}
		\caption{\retina}
	     \end{subfigure}
	    \begin{subfigure}[a]{0.32\textwidth}
           \includegraphics[width=1\textwidth]{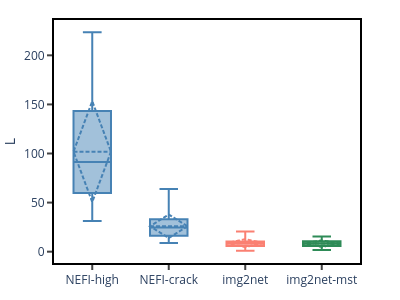}
           \caption{\pp}
    \end{subfigure}
      \caption{\textbf{Total network length}. Boxes show the distribution over the images inside each dataset of the total network length $L=\sum_{e}\ell_{e}$ calculated on the networks extracted by each method. Solid lines are the median, dashed lines are the average.}
      \label{fig:L}
\end{figure*}

We calculated the total network length as  $L=\sum_{e}\ell_{e}$ where $\ell_{e}$ is the Euclidean distance between the nodes defining the edge $e$, see Fig.~\ref{fig:L}. 
We find that \imgtnet \text{} extracts on average longer \river \text{} networks, and similar to \nefi \text{} for the \retina, but with lower variance in this case. Instead \nefi \text{} extracts much longer \pp \text{} networks, mainly due to many small minor paths permeating the whole image (this was signaled above by wider result difference in terms of similarity).   
 Instead, \mst \text{} finds smaller values in all the datasets. This highlights one important difference due to the underlying optimization setup that distinguished these two approaches. 

\begin{figure}[hptb]
    \centering
               \begin{subfigure}[a]{.5\textwidth}
         \includegraphics[width=\textwidth]{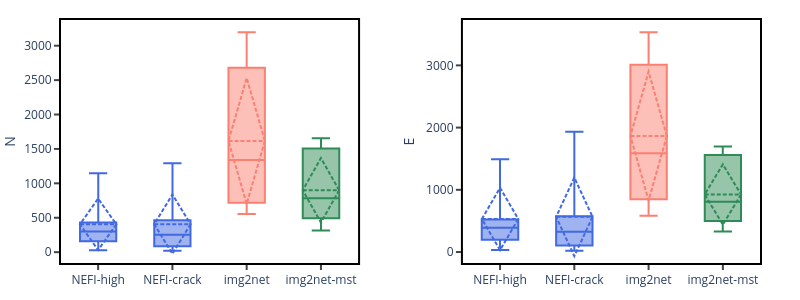}
         \caption{\river}
             \end{subfigure}
              \begin{subfigure}[a]{0.5\textwidth}
           	\includegraphics[width=\textwidth]{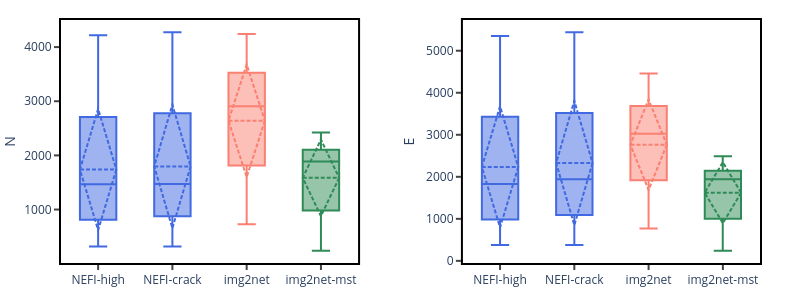}
		\caption{\retina}
	     \end{subfigure}
	    \begin{subfigure}[a]{.5\textwidth}
           \includegraphics[width=\textwidth]{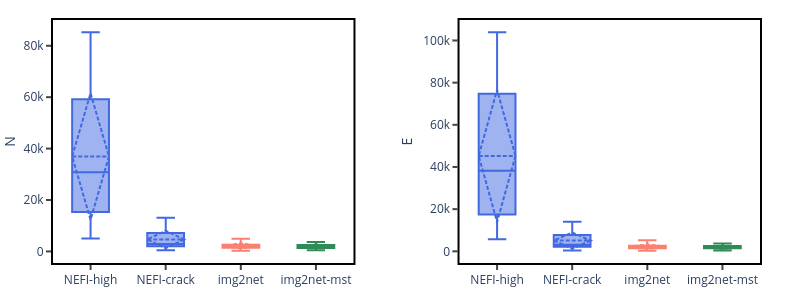}
           \caption{\pp}
    \end{subfigure}
      \caption{\textbf{Number of nodes and edges}. Boxes show the distribution over the images inside each dataset of $N$, the total number of nodes (left) and $E$, the total number of edges (right), calculated on the networks extracted by each method. Solid lines are the median, dashed lines are the average.}
      \label{fig:node-edge-plot}
\end{figure}

While a longer network total length might be due to a higher number of edges, this is not always the case. This can be seen from results on \river \text{} in Fig.~\ref{fig:node-edge-plot}, where we plot the distribution of the number of nodes and edges, other important topological properties. In fact, for these images, \nefi \text{} finds much smaller network sizes than \imgtnet, while the distribution of $L$ in Fig.~\ref{fig:L} is similar for the two routines. This is again due to \nefi \text{} representing curved parts of the network with fewer but longer straight edges, see Fig.~\ref{fig:networks-rivers-pp} for an example.
In these river networks, \imgtnet \text{} has a higher resolution, where \nefi \text{} fails to find enough details.  The opposite extreme is seen for the \pp \text{} images where \nefi \text{} has many more nodes and edges when using the routine \texttt{NEFI-high}, and also much higher $L$ as we saw before.
For the \retina \text{} vessel networks, \imgtnet \text{} extracts on average networks with higher number of nodes and edges than the other two methods, while $L$ is similar to \nefi, hence both \imgtnet \text{} and \mst \text{} have on average shorter edges than \nefi, with the difference that \imgtnet \text{} extract networks with bigger sizes.

\section{Conclusions}
We propose \imgtnet, a model for extracting networks from images. It takes as input an image and returns a network structure as a set of nodes, edges and the corresponding weights. Standard approaches for addressing these problems rely on image processing techniques. Instead, our model is based on a principled formalism adapted from recent results of optimal transport theory. We build an analogy with fluid dynamics by treating colors on pixels as fluids flowing through an image and considering a set of dynamical equations for their conductivities and flows. At convergence, these correspond to stationary solutions of a cost function that has a nice interpretation in terms of a transportation cost. 

The advantage of our approach with respect to more conventional methods is that our model naturally incorporates a principle definition of edge weights as the optimal conductivities and that can be interpreted as diameters of network edges. In addition, it allows for a principled and automatic filtering of possible redundancies by means of solving a routing optimization problem, instead of using pre-defined pruning routines.  

We test our model on various datasets,  and calculate performance measures in terms of recovering the network-like shape in the input image.
\imgtnet \text{}  performs well compared to other network extraction tools and yields networks that closely approximate the networks depicted in the images. 
In particular, it is flexible in finding various network shapes, as it can find curved geometries as those observed in river networks. 

In addition to being efficient, automated network extraction also has the advantage of yielding reproducible results and reducing human biases. Indeed, given an input image, \imgtnet \text{} will always yield the same networks, whereas manual extraction depends on the perception of the individual performing the measurement. Our model also enables practitioners to measure network-related quantities like centrality measures, branching points or curvature and angles. More importantly, given the computational efficiency of the underlying solver, it also works for large networks where manually measuring metrics across the whole network is not feasible.

In this work, we mostly show example applications from biology and ecology, but the usage of our model is not limited to this kind of networks. It can be used in a broad array of datasets to detect and measure network-like shapes. We foresee that our model will be useful for practitioners willing to perform automatic and scalable network analysis of large datasets of images.

\newpage

\section*{Acknowledgements}
The authors thank the International Max Planck Research School for Intelligent Systems (IMPRS-IS)
for supporting Diego Baptista.\\
\textbf{Author contributions:} D.B and C.D.B. derived the model, analyzed results, and wrote the manuscript. D.B. conducted the experiments.
\textbf{Competing interests}: The authors declare that they have no competing interests.
\textbf{Data and materials availability:} All data needed to evaluate the conclusions in the paper are present in the paper and/or the Supplementary Information. An open-source implementation of the code is available online at \url{https://github.com/diegoabt/Img2net}.

\bibliographystyle{ScienceAdvances}
\bibliography{bibliography}

\newcommand{\beginsupplement}{%
        \setcounter{table}{0}
        \renewcommand{\thetable}{S\arabic{table}}%
        \setcounter{figure}{0}
        \renewcommand{\thefigure}{S\arabic{figure}}%
        \setcounter{equation}{0}
        \renewcommand{\theequation}{S\arabic{equation}}
         \setcounter{section}{0}
        \renewcommand{\thesection}{S\arabic{section}}
 }

\clearpage
\beginsupplement
\begin{widetext}
\section*{{Supporting Information (SI)}}
\section{Dependency on the parameter $\delta$}
\imgtnet \text{} has two main parameters: $\delta$ and $\beta$. While $\beta$ directly acts on the optimization being performed, changing the dynamics and the transportation cost, $\delta$ controls the fine level details of the resulting conductivities and fluxes that one should keep in the final network. Specifically, the higher $\delta$, the less detailed the extracted network is. We show the impact of this in Fig.~\ref{SIfig:delta-graphs}. 

\vskip-0.5cm
\begin{figure}[h]
    \centering
    \begin{subfigure}[a]{0.99\textwidth}\label{fig:white}
        \includegraphics[width=\textwidth]{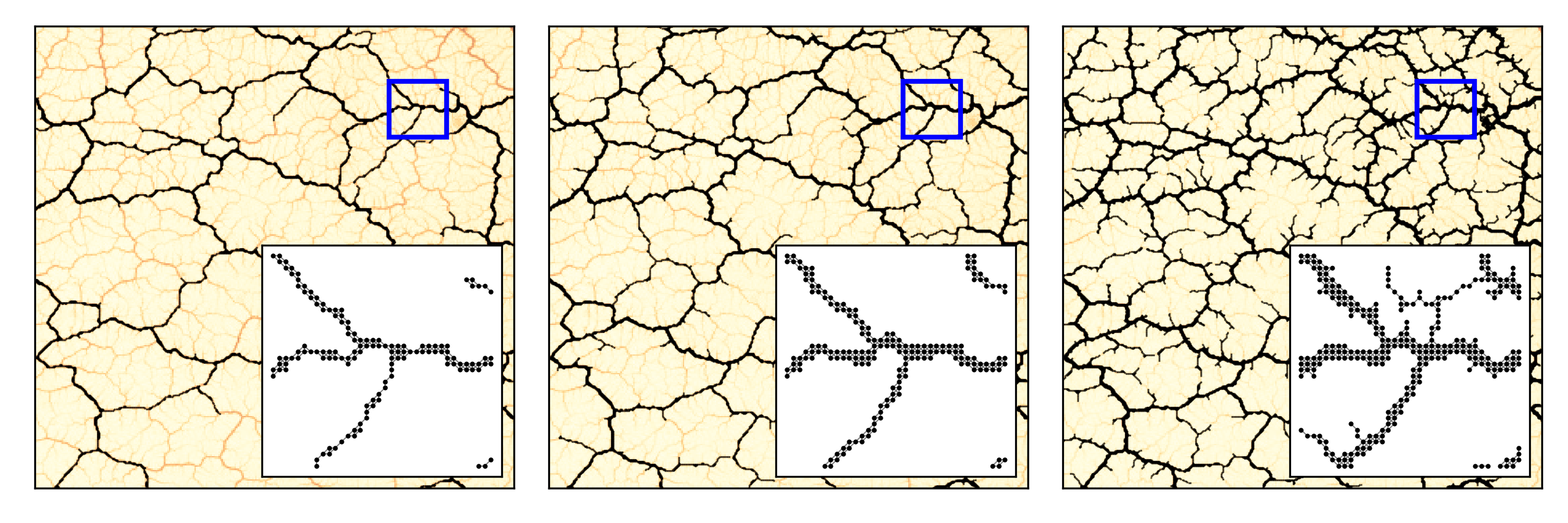}
    \end{subfigure}
      \caption{$G_{pe}$ for different values of $\delta$. The figures correspond to network extracted with $\delta=0.5, 0.4, 0.3$ (from left to right). }
      \label{SIfig:delta-graphs}
\end{figure}

\vskip-0.5cm
\section{Similarity metric}
Here we show an example of how the partition $C_\alpha$ is used to compute the similarity measure. As explained in Section III of the main manuscript, the set $[0,1]^2$ is divided into $P$ non-intersecting sets in which the difference (both binary and weighted) is computed. Fig.~\ref{SIfig:sim-mesh} shows the values of this difference inside each element of the partition. For all the experiments reported here the value of $P$ used is $(15-1)^2 = 196$, obtained by diving each axis in $15$ equally-sized intervals.
\vskip-0.4cm
\begin{figure}[h]
    \centering
    \begin{subfigure}[a]{0.99\textwidth}\label{fig:white}
        \includegraphics[width=\textwidth]{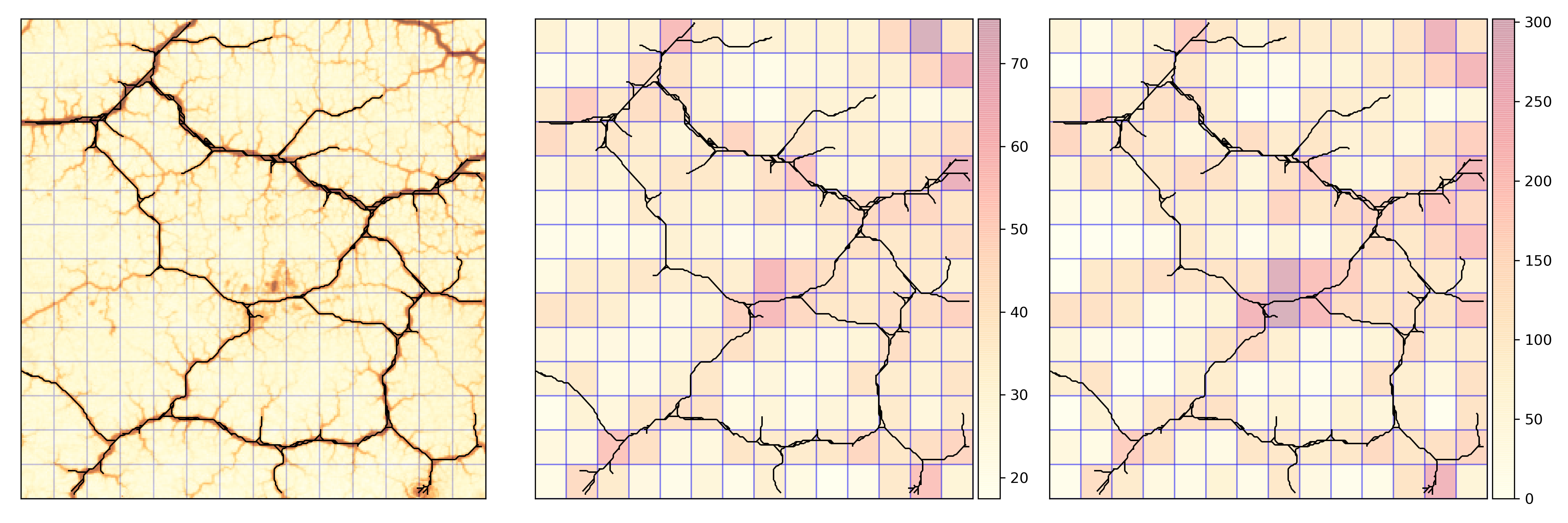}
    \end{subfigure}
      \caption{\textbf{Similarity values for a \text{Img2net} network}. Left) the input image together with the extracted network (\imgtnet) superimposed in black; center) the extracted network and the values of the weighted similarity metric $\hat{w}(G,I)$; right) the extracted network and the values of the binary similarity $\hat{w}_{b}(G,I)$. Colorbar denotes increasing values of $\hat{w}(G,I)$, $\hat{w}_{b}(G,I)$, i.e. higher color intensity signals mismatch between input image and reconstructed network, instead brighter colors signal higher similarities.   }
      \label{SIfig:sim-mesh}
\end{figure}


\newpage

\section{Dependency on $N_{runs}$}
 In Fig.~\ref{SIfig:nrunsimg} we show the impact of the parameter $N_{runs}$ on the final extracted graphs. For six example images (two per dataset), we extract the network obtained for $N_{runs} = 1,2,3,5,10,15$. The larger $N_{runs}$, the better the similarity measure, this comes at a higher computational cost which is proportional to $N_{runs}$. However, after $N_{runs}=5$ we notice that the network stabilizes to a common structure with no further main structural changes happening. 


\begin{figure}[h]
    \centering
    \begin{subfigure}[a]{0.45\textwidth}\label{fig:N-runs}
        \includegraphics[width=\textwidth]{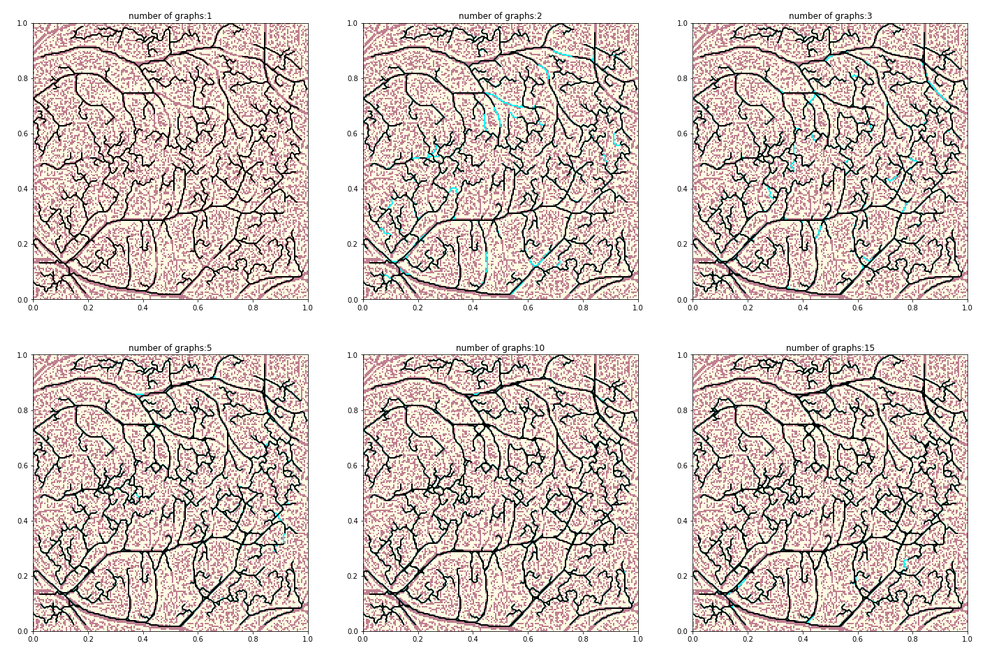}
        \caption{\retina \text{} img 0236}
    \end{subfigure}
        \begin{subfigure}[a]{0.45\textwidth}\label{fig:N-runs}
        \includegraphics[width=\textwidth]{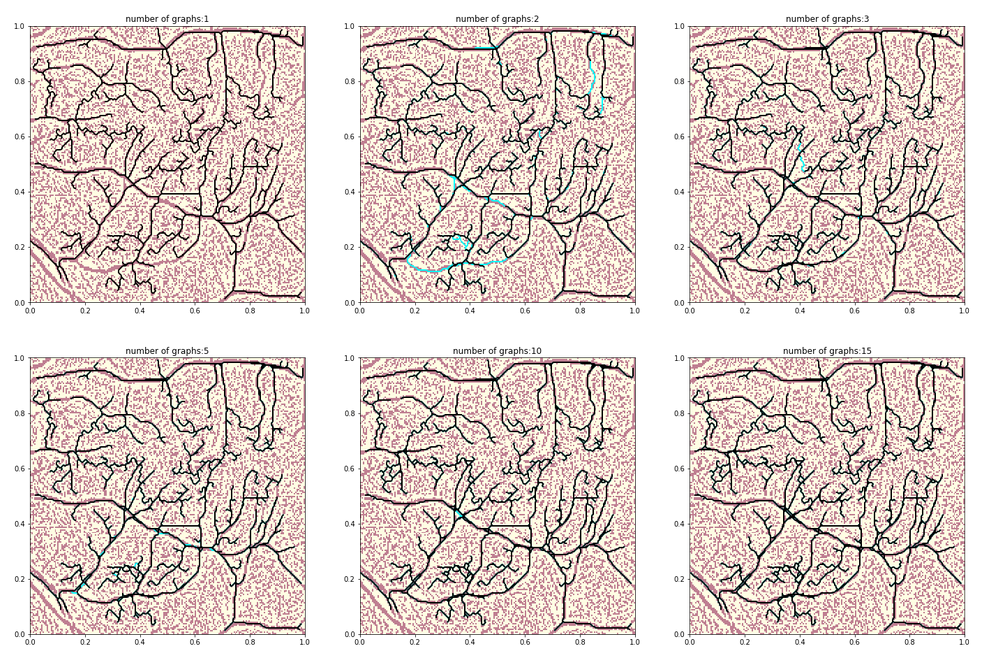}
         \caption{\retina \text{} img 0255}
    \end{subfigure}
            \begin{subfigure}[a]{0.45\textwidth}\label{fig:N-runs}
        \includegraphics[width=\textwidth]{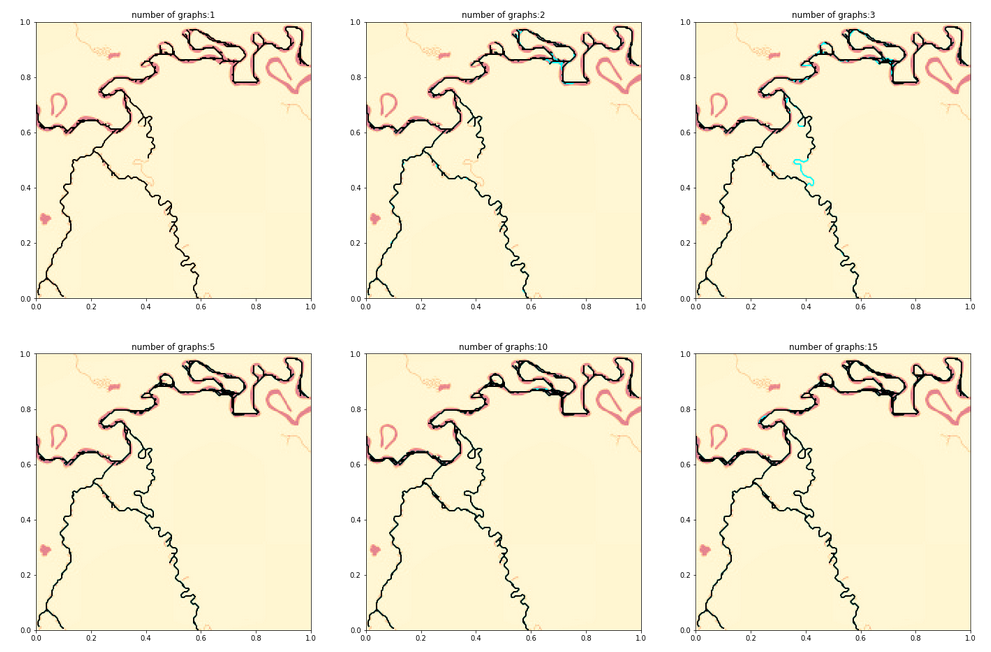}
        \caption{\river \text{} Papua river}
    \end{subfigure}
        \begin{subfigure}[a]{0.45\textwidth}\label{fig:N-runs}
        \includegraphics[width=\textwidth]{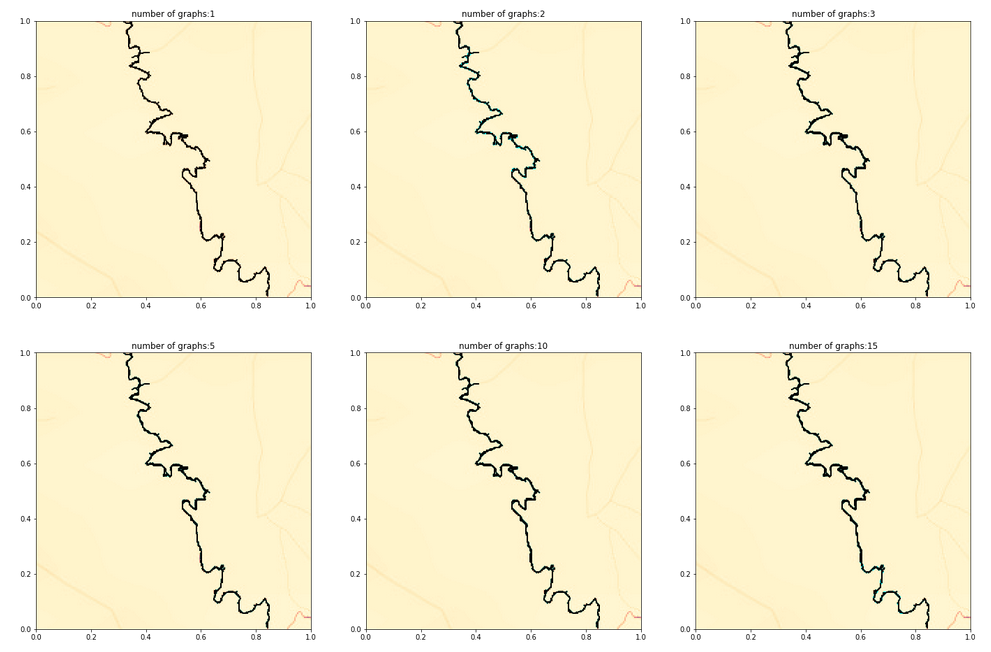}
         \caption{\river \text{} Zambian river}
    \end{subfigure}
        \begin{subfigure}[a]{0.45\textwidth}\label{fig:N-runs}
        \includegraphics[width=\textwidth]{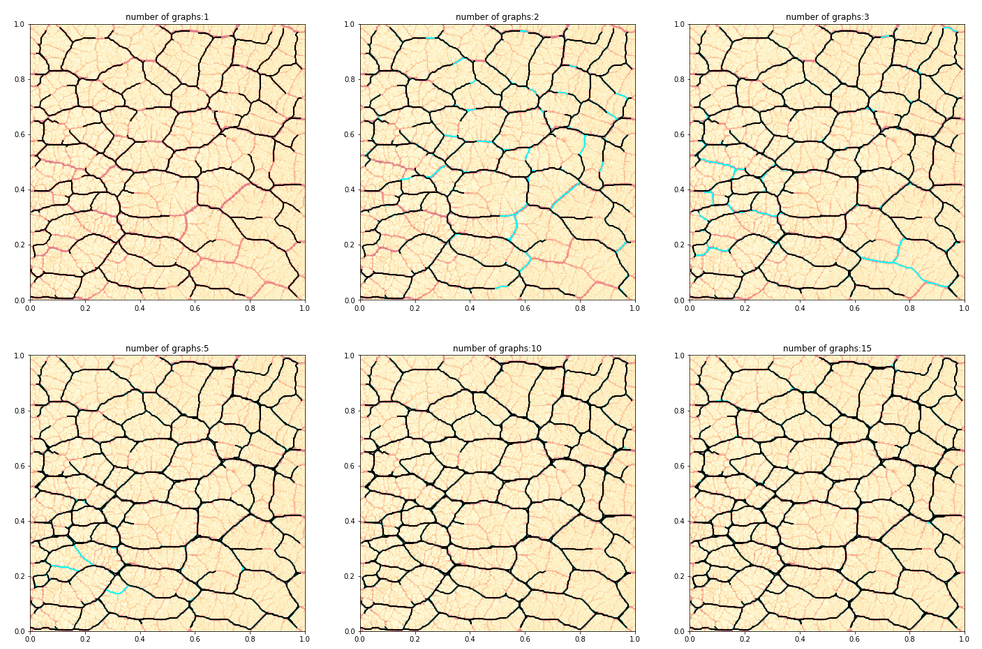}
        \caption{\pp \text{} img 0425}
    \end{subfigure}
        \begin{subfigure}[a]{0.45\textwidth}\label{fig:N-runs}
        \includegraphics[width=\textwidth]{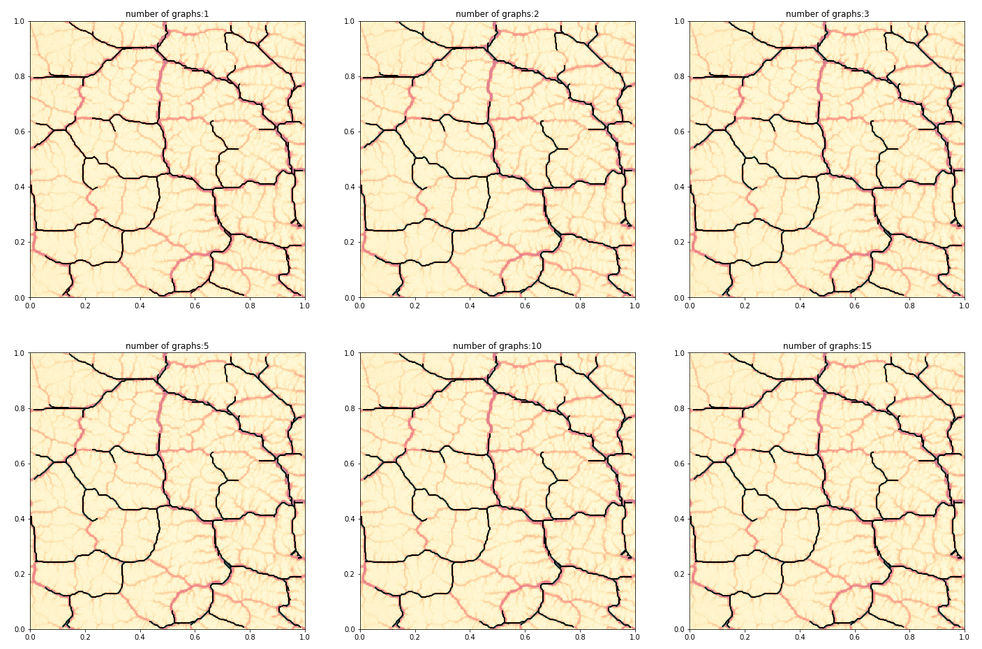}
         \caption{\pp \text{} img 1185}
    \end{subfigure}
      \caption{Impact of $N_{runs}$ in network extraction. We show six networks extracted for each input image as we vary $N_{runs}=1,2,3,5,10,15$ (indicated in the figures' titles). We highlight in light blue the edges that are newly added from the previous value of $N_{runs}$, to show the main difference between runs. The extracted networks are in black and in the background we show the input image. }
      \label{SIfig:nrunsimg}
\end{figure}

\section{\mst \text{} algorithmic details}
We give more details about the algorithmic steps behind \mst. We define $G_{mst}$, the Minimum Spanning Tree extracted graph, as the graph obtained in the following way: 

\begin{enumerate}
\item For a predefined threshold $\delta>0,$ let $G_\delta^{pe}$ be the pre-extracted graph obtained from the image $I$ as explained in Section II of the main manuscript. 
\item Get a Minimum Spanning Tree $G^{MST}_{tree}$, from $G_\delta^{pe}$ using
\begin{verbatim}
networkx.algorithms.
tree.maximum_spanning_edges(G_pe,
  algorithm='kruskal', data='weight')
 \end{verbatim}
\item Compute $\mathcal{L}(G^{MST}_{tree})$,  the set of leaves of $G^{MST}_{tree}$.
\item Define the set of \textit{terminals} $T$ by taking a random subset of $\mathcal{L}(G^{MST}_{tree})$ of predefined size (in our experiments $|T| = 0.025 \times |\mathcal{L}(G^{MST}_{tree})|,$ i.e. we take a random set $T$ of size equal to $2.5\%$ of the total number of leaves). 
\item Compute an approximation of the Steiner tree (see definition below) from $G_\delta^{pe}$ using $T$ as terminal set by calling the function
\begin{verbatim}
networkx.algorithms.approximation.
steiner_tree(G_pe, T)
 \end{verbatim}

 Repeat this step for a predefined number $N_{runs}$ of times, to get a family of steiner trees $\mathcal{F}=\{G^{st}_1,G^{st}_2, ..., G^{st}_{N_{runs}}\},$ built from $G_\delta^{pe}$ using different random sets $T$ of the same size.
\item Superimpose all the elements of $\mathcal{F}$ to get a graph $G_{mst}=(V,E,W),$ s.t. $V,E$ and $W$ are defined as in Section II Eq. (6) in the main manuscript. 
\end{enumerate}

Given a weighted undirected graph $G$ and a set of terminals $T\subset V(G),$ the Steiner Tree problem consists on finding a subgraph $G_{st}$ of $G$ s.t. $T\subset V(G_{st}),$ and 
$$
G_{st}\in \underset{S\subset G, S \text{ tree}}{\text{arg min}} \sum_{e\in E(S)} \! \! w(e),
$$
where $S\subset G$ means that $S$ is a subgraph of $G$, and $w(e)\geq 0$ represents the weight of the edge $e$ on $G$. 

\vspace{1cm}

\section{Details about the input images}

The images used for the experiments can be found as:
\begin{enumerate}
\item \retina : these are the images provided on \cite{hoover2000locatingbv} (20 images in total).
\item \river : these are screenshots taken from \cite{openseamap}.  Each image has an associated permanent link. All the permanent links are stored in the repository.
\item \pp : the images are taken from \cite{dirnberger2017introducing}. These images can be found in the website in different subfolders named \textit{motions} inside the \textit{processed\_images} folder. Namely, the 25 used images are: \textit{IMG\_0049, IMG\_0335  IMG\_0486,  IMG\_0861,  IMG\_1162,
IMG\_0263,  IMG\_0399,  IMG\_0750,  IMG\_0874,  IMG\_1185,
IMG\_0281,  IMG\_0409,  IMG\_0756,  IMG\_0961,  IMG\_1231,
IMG\_0286,  IMG\_0425,  IMG\_0797,  IMG\_0968,  IMG\_1270,
IMG\_0303,  IMG\_0430,  IMG\_0810,  IMG\_1039,  IMG\_1621}.
\end{enumerate}

\vspace{.5cm}

\section{Image preprocessing} \label{SIsec:imgpre}

This section contains the information related to the steps previous to the network extraction. We preprocessed the images by applying
 color enhancing, grayscale mapping, segmentation, cropping, and resizing.

\subsection{Cropping}

All of the images $I$ were cropped in order to make them square shaped (\textit{width = height}). We obtained the cropped image by specifying the \textit{center} of the cropping \texttt{(x,y)} and the \textit{width} \texttt{w}, and then using \textit{OpenCV (cv)} functionalities
\\
\begin{verbatim}
w = int(min(x,y)/3)-1
x = int(x/2)
y = int(y/2)  
img = cv.imread(image) 
crop_img = img[x-w:x+w, y-w:y+w]
\end{verbatim}

\subsection{Grayscale mapping}

For the \retina {} and \river {} datasets we converted the colorful cropped images into grayscales by using
\\
\begin{verbatim}
gray = cv.cvtColor(img,cv.COLOR_BGR2GRAY)
\end{verbatim}

\subsection{Color enhancing}
For the \retina {} and \river {} datasets we converted the grayscale images into enhanced ones by using
\\
\begin{verbatim}
clahe = cv.createCLAHE(clipLimit=3.0,
        tileGridSize=(8,8))
img = clahe.apply(gray)
\end{verbatim}

\subsection{Segmentation}
For the \retina {} dataset we segmented the enhanced images by applying
\\
\begin{verbatim}
img = cv.adaptiveThreshold(img,
      255,cv.ADAPTIVE_THRESH_GAUSSIAN_C,
      cv.THRESH_BINARY,11,2)
\end{verbatim}

\subsection{Resizing}

The procedures proposed in this paper depend on the size of the input images. Thus, before extracting the networks, images $I$ are mapped into smaller ones by using a procedure called \textit{resizing\_image()}. The idea behind this function is to get a \textit{similar representation} of $I$ by replacing it by other image $I'$ in which pixel values are a function of the values of the pixels in $I$. The functions used to generate the new values are the ones accepted by the \textit{OpenCV} function \textit{resize} (input \textit{interpolation}). We set \textit{interpolation} to be either \textit{cv.INTER\_NEAREST} or \textit{cv.INTER\_AREA} for all the experiments exhibited in this work. The former does nearest-neighbour interpolation while the latter assigns the new pixel value as the \textit{area} of the surrounding pixels in a given neighborhood. We used \textit{cv.INTER\_AREA} for \pp {} and \river, and \textit{cv.INTER\_NEAREST} for \retina. Images in \pp, \river {} and \retina {} were reduced to have 300, 200 and 200 width, respectively.


\end{widetext}

\end{document}